\documentclass[sigconf, nonacm]{acmart}
\AtBeginDocument{%
  \providecommand\BibTeX{{\rm B\kern-.05em{\sc i\kern-.025em b}\kern-.08em\TeX}}}

\setcopyright{none}
\renewcommand\footnotetextcopyrightpermission[1]{} 
\usepackage{bm} 
\usepackage{mathtools}
\usepackage{booktabs}
\usepackage{graphicx}
\usepackage{adjustbox}
\usepackage{multirow}
\usepackage{enumitem}
\usepackage{xspace}
\usepackage{microtype}
\usepackage{xcolor}
\usepackage{cleveref}
\usepackage{subcaption}
\usepackage{algorithm}
\usepackage{algpseudocode}
\usepackage{tcolorbox}

\usepackage{tabularx}
\usepackage{array}
\usepackage{ragged2e}
\usepackage{fontawesome5}
\newcolumntype{Y}{>{\RaggedRight\arraybackslash}X}
\newcolumntype{P}[1]{>{\RaggedRight\arraybackslash}p{#1}}

\hypersetup{citecolor=cyan}
\crefformat{section}{\S#2\color{blue}#1#3} 
\crefformat{subsection}{\S#2\color{blue}#1#3}
\crefformat{subsubsection}{\S#2#1#3}
\crefname{equation}{Eq.}{Eqs.}

\def\eqref#1{equation~\ref{#1}}

\def\1{\bm{1}}

\DeclareMathAlphabet{\mathsfit}{\encodingdefault}{\sfdefault}{m}{sl}
\SetMathAlphabet{\mathsfit}{bold}{\encodingdefault}{\sfdefault}{bx}{n}

\newcommand{\softmax}{\mathrm{softmax}}

\newcommand{\modelname}{ChronoSSM\xspace}
\newcommand{\joint}{\textsc{Joint}\xspace}
\newcommand{\twostage}{\textsc{Two-Stage}\xspace}

\newcommand{\parab}[1]{\smallskip\noindent {\bf #1}}

\title{\modelname: Training for Temporally Aware Representations in Autoregressive State Space Models}

\author{Adrien Schoen}
\authornote{Both authors contributed equally to this research.}
\email{adrien.schoen@cnrs.fr}
\affiliation{
  \institution{ENS de Lyon, CNRS, UCBL1, LIP}
  \city{Lyon}
  \country{France}
}

\author{Nachiketa Ratnakar Patil}
\authornotemark[1]
\email{npatil@iitb.ac.in}
\affiliation{
  \institution{Centre for Machine Intelligence and Data Science}
  \institution{Indian Institute of Technology Bombay}
  \city{Mumbai}
  \country{India}
}

\author{Arjun Bhagoji}
\email{arjunp@iitb.ac.in}
\affiliation{
  \institution{Centre for Machine Intelligence and Data Science}
  \institution{Indian Institute of Technology Bombay}
  \city{Mumbai}
  \country{India}
}

\author{Francesco Bronzino}
\email{francesco.bronzino@ens-lyon.fr}
\affiliation{
  \institution{ENS de Lyon, CNRS, UCBL1, LIP}
  \city{Lyon}
  \country{France}
}
\affiliation{
  \institution{Institut universitaire de France}
  \country{France}
}

\begin{document}

\begin{abstract}
Modern sequence models, from Transformers to State Space Models, have enabled powerful generative modeling across diverse domains, yet they are typically trained to predict \emph{what} happens while treating \emph{when} it happens as a secondary concern. In data-mining settings where events are associated with explicit timing information, this separation can limit temporal reasoning, anomaly detection, and faithful reconstruction of event chronology. A common strategy is to treat timing as an auxiliary signal, training a separate timing model using representations learned solely for event prediction. However, this two-stage approach implicitly assumes that representations optimized for event prediction already contain sufficient temporal structure.

We introduce \modelname{}, an autoregressive State Space Model (SSM) that jointly models events and timestamps with a shared backbone trained using combined token and temporal generation objectives. We compare the \joint regime, where temporal supervision updates the backbone, with the \twostage regime, where timing is learned only using the frozen event representations. Across four domains spanning dense and partial timestamp supervision, \joint training consistently makes inter-arrival information more recoverable from frozen representations without any systematic degradation in content-generation quality overall. Our results show that temporal supervision can produce more temporally informative representations without materially degrading autoregressive event modeling.
\end{abstract}

\maketitle
\fancyfoot[L]{\textit{Preprint}}

\section{Introduction}
\label{sec:introduction}
Modern autoregressive sequence models are now used well beyond natural language, including in domains such as business-process traces, clinical event sequences derived from electronic health records, network traffic, and temporal knowledge graphs~\cite{vandongen2018bpi,johnson2023mimic,networkdataset,renet}. In these settings, events are associated with explicit temporal metadata, and downstream tasks depend not only on what happened but also on when it happened~\cite{lanvin,fraud_detection,flowchronicle,renet}. This raises a basic question: \emph{can autoregressive sequence models learn both event identity and event timing within a shared representation}? We formalize this problem in \cref{sec:context}.

A common way to incorporate timing has been to treat it as an auxiliary task~\cite{jiang2024netdiffusion,chu2025netssm}: first train a sequence model for next-event prediction, then freeze its representations and fit a separate timing module on top. This two-stage strategy is simple and modular, but it assumes that representations optimized only for token prediction already retain enough temporal information. That assumption is questionable: the token generation objective only rewards features that improve next-event likelihood, not inter-arrival modeling. A downstream timing module must therefore recover temporal structure from representations that were never explicitly trained to encode it.

In this paper, we introduce \modelname{}~(\cref{sec:method}), an autoregressive State Space Model that predicts both the next event and its timestamp from a shared backbone. A lightweight temporal head is trained jointly with the token prediction head, so temporal gradients can directly update the backbone. The resulting model learns event semantics and temporal dynamics within a unified representation.
Our code and configurations shall be released upon acceptance.

To test whether this joint formulation offers a real advantage over the two-stage alternative, we conduct a controlled comparison between two training regimes that share the same architecture, data, and temporal objective (\cref{sec:experimental-setup}): a \twostage regime, in which timing is learned after event prediction on frozen representations, and a \joint regime, in which both objectives are optimized together. We evaluate this comparison on four key domains spanning dense and partial timestamp supervision: business-process traces~\cite{vandongen2018bpi,mannhardt2018bpi}, clinical event sequences~\cite{johnson2023mimic}, network traffic~\cite{networkdataset,chu2025netssm}, and temporal knowledge graphs~\cite{renet}. In addition to these four main domains, we also experiment with symbolic-music~\cite{kong2022giantmidi,hawthorne2019maestro,huang2019musictransformer} generation reported in \cref{app:music-experiments}. Our two main research questions are:

\parab{RQ1 (Temporal Recoverability).} Does \joint training yield representations from which inter-arrival information is more recoverable than under \twostage training? We evaluate this using a suite of \emph{recoverability diagnostics} (formalized in \cref{sec:recoverability}) that probe whether temporal structure is encoded in frozen learned representations and can be extracted by lightweight analyses. We find that \joint training consistently produces representations from which timing is more recoverable across all four domains, with the magnitude of the effect varying by domain.

\parab{RQ2 (Quality Degradation).} Does injecting temporal supervision into the shared backbone systematically degrade downstream modeling quality? Stronger temporal recoverability is only useful if it does not compromise the model's primary event-generation function. We therefore compare \joint{} and \twostage{} training on domain-specific generation metrics and find no evidence of a systematic degradation under \joint{} training (\cref{sec:results}), with \joint{} generating higher quality data in 2 out of 4 domains.

Overall, these results indicate that explicit temporal supervision can improve temporal recoverability without systematically degrading the model's primary generative function. This paves the way for temporally-aware generative models to be trained and deployed in relevant domains.
\section{Background}\label{sec:context}

This section introduces the modeling framework used throughout the paper. We first describe the underlying sequence model and the representations it produces, and then formalize the problem setting for predicting the next event and its associated timing information.

\subsection{The State-Space Architecture}

State Space Models (SSMs) are sequence models with recurrent latent-state dynamics that support linear-time processing in sequence length~\cite{ssm-better,ssm-paper}. This is useful in long-sequence settings, where computational efficiency becomes important as context length grows.

The core mechanism of an SSM is a linear dynamical system that maps an input sequence to an output sequence through a hidden state~\cite{ssm-paper}. At step $k$, the model updates a latent state $s_k$ from the previous state and the current input $x_k$, and then produces an output representation $h_k$:
\begin{align}
  s_k &= A_k s_{k-1} + B_k x_k, \\
  h_k &= C_k s_k + D_k x_k.
\end{align}
In modern selective SSMs, some of these parameters are conditioned on the current input, allowing the model to selectively propagate or discard information according to the sequence context. This input-dependent selection mechanism enables the model to adapt its effective memory to the content of the sequence.

Stacking several SSM layers produces a causal sequence model that processes the input in a single forward pass and generates a hidden representation $h_k$ at each position. These token-wise hidden states can then be used by lightweight prediction heads for tasks such as next-token prediction or temporal prediction. This representation-level property is the one exploited in this paper.

\subsection{Problem Setting}\label{ss:problem_formulation}

Let $\mathcal{D}$ denote a dataset of timestamped sequences. Each sequence of length $T$ in $\mathcal{D}$ is written as $\{(x_k,t_k)\}_{k=0}^{T-1}$, where $x_k$ is a discrete token and $t_k$ is its associated timestamp. From successive timestamps, we derive the inter-arrival times $\{\Delta t_k=t_k-t_{k-1}\}_{k=1}^{T-1}$.

Denoting the token history up to step $k$ by $\mathcal{X}_k=\{x_i\}_{i=0}^{k}$, our objective is to model the joint distribution of the next token $x_{k+1}$ and its associated inter-arrival time
$\Delta t_{k+1}$:
\begin{equation}
  p_{\gamma}
  (x_{k+1},\Delta t_{k+1}\mid\mathcal{X}_k).
  \label{eq:iat}
\end{equation}

Using the chain rule, we factor this joint distribution as
\begin{equation}
  p_{\gamma}
  (x_{k+1},\Delta t_{k+1}\mid\mathcal{X}_k)
  =
  p_{\theta,\alpha}
  (x_{k+1}\mid\mathcal{X}_k)\,
  p_{\theta,\phi}
  (\Delta t_{k+1}\mid x_{k+1},\mathcal{X}_k),
  \label{eq:joint-factorization}
\end{equation}
where $\gamma=\{\theta,\alpha,\phi\}$ collects the parameters of the shared sequence model, token-prediction head, and temporal-prediction head, respectively. Let $V$ denote the vocabulary size and $D$ the hidden-state dimension.

The shared sequence model is a causal autoregressive model parameterized by $\theta$ that produces a token-wise hidden representation $h_k$ from the history $\mathcal{X}_k$. In our main instantiation, we use a State Space Model, although the framework is not tied to this particular architecture. Any causal autoregressive backbone that produces token-wise hidden representations can be used instead; Appendix~\cref{app:backbone-robustness} illustrates this with an additional Transformer-based instantiation. Formally, the backbone computes
\begin{equation}
  h_k
  =
  \mathrm{Backbone}_{\theta}(\mathcal{X}_k).
  \label{eq:backbone-hidden-state}
\end{equation}

The token-prediction head is a linear map $W^{\mathrm{tok}}_{\alpha}\in\mathbb{R}^{V\times D}$ parameterized by $\alpha$, and it models the next-token distribution as
\begin{equation}
  p_{\theta,\alpha}(x_{k+1}\mid\mathcal{X}_k)
  =
  \softmax
  \left(
    W^{\mathrm{tok}}_{\alpha}h_k
  \right).
  \label{eq:next-token-dist}
\end{equation}

The temporal-prediction head is a function $f^{\mathrm{time}}_{\phi}(\cdot; r_{k+1})$ parameterized by $\phi$ and conditioned on a transition representation $r_{k+1}$. After $x_{k+1}$ is processed, the sequence model produces $h_{k+1}$ according to \cref{eq:backbone-hidden-state}. Because $\Delta t_{k+1}$ describes the transition from position $k$ to position $k+1$, its prediction is based on a representation $r_{k+1}$ derived from both $h_k$ and $h_{k+1}$. The temporal-prediction head models the second factor as
\begin{equation}
  p_{\theta,\phi}
  (\Delta t_{k+1}\mid x_{k+1},\mathcal{X}_k)
  =
  f^{\mathrm{time}}_{\phi}
  \left(
    \Delta t_{k+1}; r_{k+1}
  \right).
  \label{eq:temporal-factor}
\end{equation}
The construction of $r_{k+1}$ is introduced in \cref{sec:method}.

\section{\modelname}\label{sec:method}

Having defined the modeling setting and notation, we now describe the design choices that instantiate \modelname{}. We first introduce the temporal representation $r_{k+1}$ supplied to the temporal-prediction head in \cref{eq:temporal-factor}. We then define the training objective and compare two optimization schedules that differ in whether temporal supervision is allowed to update the shared backbone.

\subsection{Model Design}
\label{ss:model_design}
An inter-arrival time measures the elapsed time between two consecutive tokens, rather than a property of either token considered in isolation. We therefore represent it using the change between their corresponding hidden states:
\begin{equation}
  r_{k+1}
  =
  \Delta h_{k+1}
  =
  h_{k+1}-h_k.
  \label{eq:hidden-transition}
\end{equation}

The temporal target and the representation used to predict it are thus both defined as differences between consecutive positions:
\begin{equation}
  \underbrace{\Delta t_{k+1}=t_{k+1}-t_k}_{\text{temporal difference}}
  \qquad\longleftrightarrow\qquad
  \underbrace{\Delta h_{k+1}=h_{k+1}-h_k}_{\text{hidden-state difference}}.
\end{equation}
This choice for $r_{k+1}$ is a modeling decision rather than a structural requirement of the framework.
An alternative temporal representation $r'_{k+1} = [h_{k}; h_{k+1}]$ is considered in
Appendix~\cref{app:temporal-representations}.

\subsection{Training Objective}

Under the factorization in \cref{eq:joint-factorization}, maximum-likelihood training decomposes the negative log-likelihood of each transition into a token-prediction term and a temporal-prediction term:
\begin{align}
  \ell_k
  &=
  -\log p_{\gamma}
  (x_{k+1},\Delta t_{k+1}\mid\mathcal{X}_k)
  \nonumber\\
  &=
  \underbrace{
    -\log p_{\theta,\alpha}
    (x_{k+1}\mid\mathcal{X}_k)
  }_{\ell_k^{\mathrm{tok}}}
  +
  \underbrace{
    -\log p_{\theta,\phi}
    (\Delta t_{k+1}\mid x_{k+1},\mathcal{X}_k)
  }_{\ell_k^{\mathrm{time}}}.
  \label{eq:per-step-nll}
\end{align}
We now instantiate these two terms using the token-prediction and temporal-prediction heads introduced in \cref{ss:problem_formulation}.

For token prediction, the factor $p_{\theta,\alpha}(x_{k+1}\mid\mathcal{X}_k)$ is categorical, so the sequence-level token loss is
\begin{equation}
  L_{\mathrm{tok}}
  =
  \sum_{k=0}^{T-2}
  \ell_k^{\mathrm{tok}}
  =
  \sum_{k=0}^{T-2}
  \mathrm{CE}
  \left(
    W^{\mathrm{tok}}_{\alpha}h_k,
    x_{k+1}
  \right),
  \label{eq:token-loss}
\end{equation}
where $\mathrm{CE}$ denotes categorical cross-entropy, which is appropriate because $\{x_k\}_{k=0}^{T-1}$ is a sequence of discrete tokens~\cite{goodfellow_dl_2016} (see \cref{ss:problem_formulation}).

For temporal prediction, the factor $p_{\theta,\phi}(\Delta t_{k+1}\mid x_{k+1},\mathcal{X}_k)$ is defined by a conditional density over the continuous-valued target $\Delta t_{k+1}$. This definition is well posed provided that the chosen conditional density has support on the observed inter-arrival targets, including zero when zero inter-arrival times are present. We therefore define the sequence-level temporal loss as the negative log-likelihood of the observed inter-arrival times under that density:
\begin{equation}
  L_{\mathrm{time}}
  =
  \sum_{k=0}^{T-2}
  \ell_k^{\mathrm{time}}
  =
  \sum_{k=0}^{T-2}
  -\log
  f^{\mathrm{time}}_{\phi}
  \left(
    \Delta t_{k+1};
    \Delta h_{k+1}
  \right),
  \label{eq:time-loss}
\end{equation}
where $\Delta h_{k+1}$ is the hidden-state transition defined in \cref{eq:hidden-transition}. The concrete form of $f^{\mathrm{time}}_\phi$ is specified for each experimental regime in \cref{sec:experimental-setup}.

Summing \cref{eq:per-step-nll} over the sequence gives the joint negative log-likelihood
\begin{equation}
  \mathcal{L}_{\mathrm{NLL}}
  =
  L_{\mathrm{tok}}+L_{\mathrm{time}}.
  \label{eq:joint-nll}
\end{equation}
Minimizing this objective corresponds to maximum-likelihood training of the factorized distribution over tokens and inter-arrival times.

In practice, the token and temporal losses may have different numerical scales. We therefore optimize the weighted objective
\begin{equation}
  \mathcal{L}
  =
  L_{\mathrm{tok}}
  +
  \lambda_{\tau}L_{\mathrm{time}},
  \label{eq:joint-objective}
\end{equation}
where $\lambda_{\tau}\geq0$ controls the relative contribution of temporal prediction.

\subsection{Training Schedules}
\label{ss:train_schedules}
\begin{figure*}[t]
\centering

\begin{minipage}[t]{0.48\textwidth}
    \vspace{0pt}
    \centering

    \begin{subfigure}[t]{\linewidth}
        \centering
        \includegraphics[width=0.95\linewidth]
        {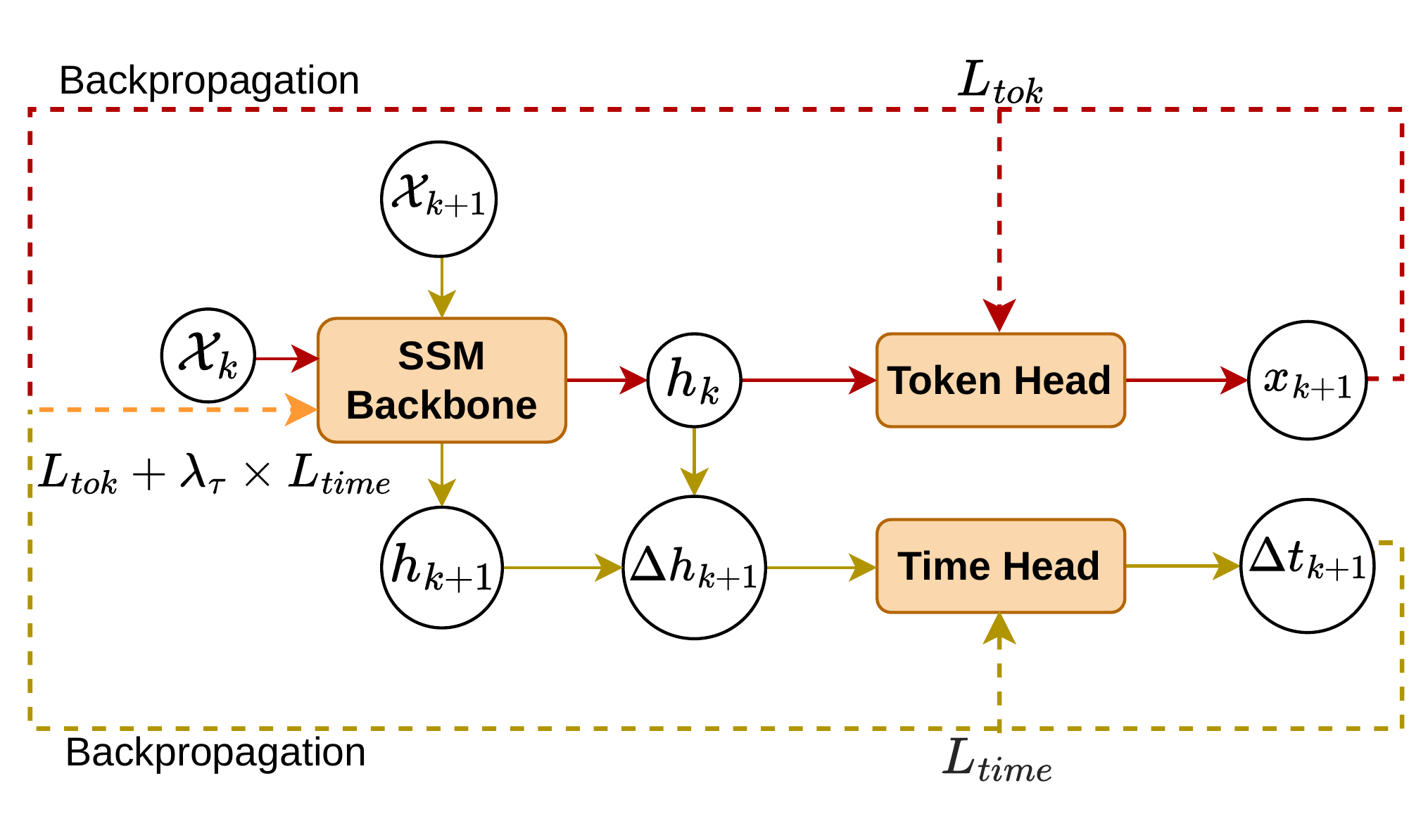}
        \caption{\joint{} training updates the shared SSM backbone and both
        prediction heads using the token and temporal losses.}
        \label{fig:training_joint}
    \end{subfigure}

    \vspace{0.8em}

    \begin{subfigure}[t]{\linewidth}
        \centering
        \includegraphics[width=0.95\linewidth]
        {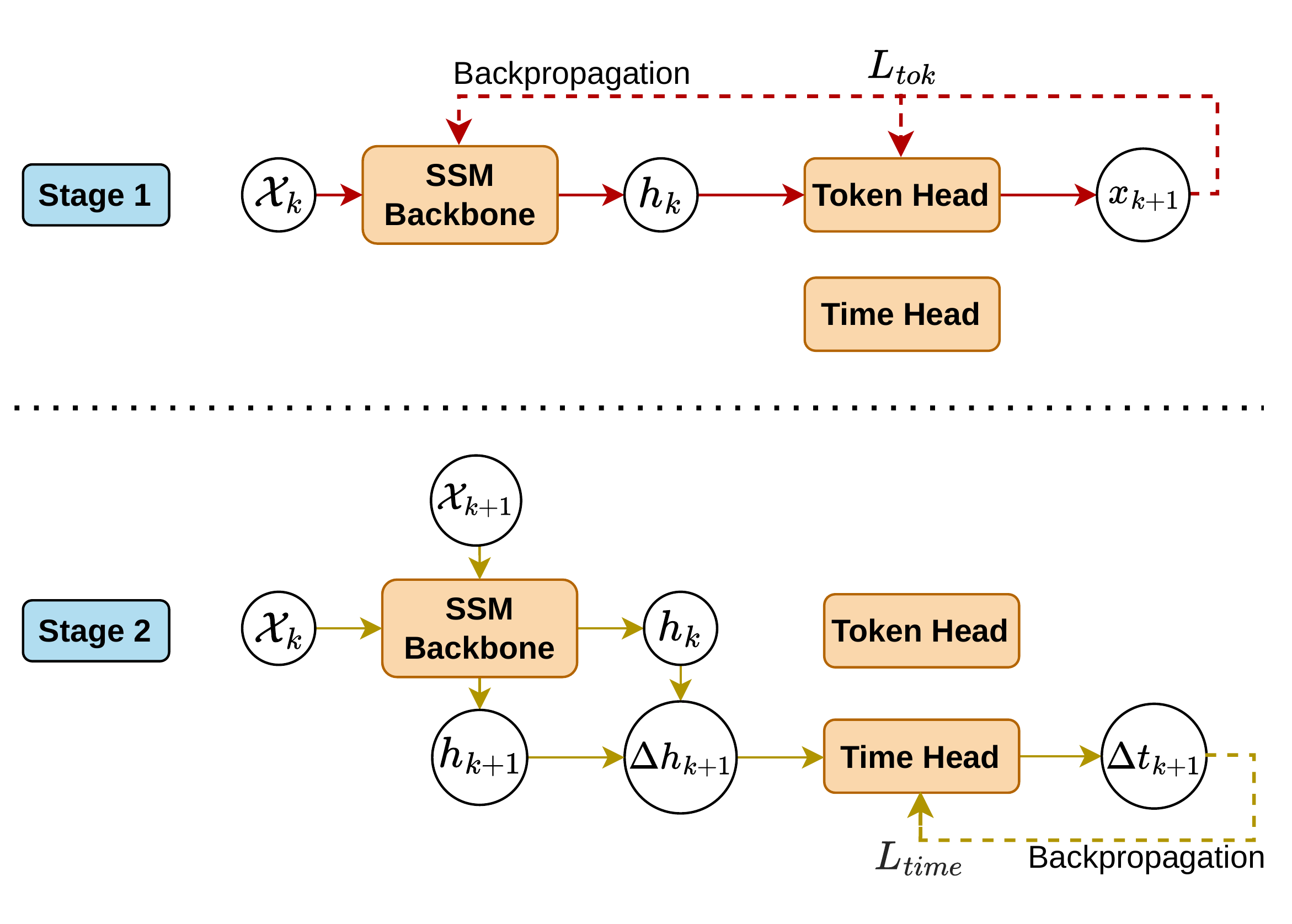}
        \caption{\twostage{} training first optimizes the shared SSM backbone
        and token-prediction head, then trains the temporal-prediction head
        in a separate second stage.}
        \label{fig:training_twostage}
    \end{subfigure}

    \caption{Overview of the \joint{} and \twostage{} training regimes.}
    \label{fig:training}
\end{minipage}
\hfill
\begin{minipage}[t]{0.48\textwidth}
    \vspace{0pt}

    \begin{algorithm}[H]
      \caption{\joint{} training}
      \label{alg:training-joint}
      \begin{algorithmic}[1]
        \Require Epochs $E$; backbone $\mathrm{Backbone}_{\theta}$;
        token head $W^{\mathrm{tok}}_{\alpha}$;
        temporal head $f^{\mathrm{time}}_{\phi}$;
        dataset $\mathcal{D}$ of timestamped sequences;
        sequence length $T$; temporal weight $\lambda_{\tau}$
        \For{epoch $=1,\ldots,E$}
          \For{each $\{(x_k,t_k)\}_{k=0}^{T-1}\in\mathcal{D}$}
            \State $(h_0,\ldots,h_{T-1})
            \gets\mathrm{Backbone}_{\theta}(x_0,\ldots,x_{T-1})$
            \State Compute $L_{\mathrm{tok}}$ using
            \cref{eq:token-loss}
            \State Compute $L_{\mathrm{time}}$ using
            \cref{eq:time-loss}
            \State $\mathcal{L}
            \gets L_{\mathrm{tok}}
            +\lambda_{\tau}L_{\mathrm{time}}$
            \State Update $\alpha$ using $\nabla L_{\mathrm{tok}}$
            \State Update $\phi$ using $\nabla L_{\mathrm{time}}$
            \State Update $\theta$ using $\nabla\mathcal{L}$
          \EndFor
        \EndFor
      \end{algorithmic}
    \end{algorithm}

    \vspace{-2em}

    \begin{algorithm}[H]
      \caption{\twostage{} training}
      \label{alg:training-twostage}
      \begin{algorithmic}[1]
        \Require Token-training epochs $E_{\mathrm{tok}}$;
        temporal-training epochs $E_{\mathrm{time}}$;
        backbone $\mathrm{Backbone}_{\theta}$;
        token head $W^{\mathrm{tok}}_{\alpha}$;
        temporal head $f^{\mathrm{time}}_{\phi}$;
        dataset $\mathcal{D}$ of timestamped sequences;
        sequence length $T$

        \For{epoch $=1,\ldots,E_{\mathrm{tok}}$}
          \For{each $\{(x_k,t_k)\}_{k=0}^{T-1}\in\mathcal{D}$}
            \State $(h_0,\ldots,h_{T-1})
            \gets\mathrm{Backbone}_{\theta}(x_0,\ldots,x_{T-1})$
            \State Compute $L_{\mathrm{tok}}$ using
            \cref{eq:token-loss}
            \State Update $\theta$ and $\alpha$ using
            $\nabla L_{\mathrm{tok}}$
          \EndFor
        \EndFor

        \State Freeze $\theta$ and $\alpha$

        \For{epoch $=1,\ldots,E_{\mathrm{time}}$}
          \For{each $\{(x_k,t_k)\}_{k=0}^{T-1}\in\mathcal{D}$}
            \State $(h_0,\ldots,h_{T-1})
            \gets\mathrm{Backbone}_{\theta}(x_0,\ldots,x_{T-1})$
            \State Compute $L_{\mathrm{time}}$ using
            \cref{eq:time-loss}
            \State Update $\phi$ using
            $\nabla L_{\mathrm{time}}$
          \EndFor
        \EndFor
      \end{algorithmic}
    \end{algorithm}

\end{minipage}

\end{figure*}

We consider two approaches for optimizing the parameter set $\gamma=\{\theta,\alpha,\phi\}$. Both regimes use the same shared sequence backbone model, token-prediction head, temporal-prediction head, and loss components. They differ only in whether the temporal loss is allowed to update the shared backbone. The two regimes are illustrated in \Cref{fig:training}: \Cref{fig:training_joint} shows the \joint{} setup, whereas \Cref{fig:training_twostage} shows the \twostage{} schedule. The corresponding optimization procedures are summarized in \Cref{alg:training-joint,alg:training-twostage}.

\parab{\joint{} Training.}
Under \joint{} training, the backbone and both prediction heads are optimized simultaneously using the objective in \cref{eq:joint-objective}. Gradients from both $L_{\mathrm{tok}}$ and $L_{\mathrm{time}}$ update the backbone parameters $\theta$, while the token-prediction and temporal-prediction heads are updated by $L_{\mathrm{tok}}$ and $L_{\mathrm{time}}$, respectively.

In particular, the temporal loss backpropagates through $\Delta h_{k+1}=h_{k+1}-h_k$, encouraging changes between consecutive hidden states to become informative about the corresponding inter-arrival times. Because the token-prediction head uses hidden states produced by the same backbone, temporal supervision can also affect the representations used for next-token prediction.

\parab{\twostage{} Training.}
Under \twostage{} training, token and temporal learning are separated into two phases. In the first phase, the backbone and token-prediction head are optimized using $L_{\mathrm{tok}}$ alone. In the second phase, the temporal-prediction head is trained using $L_{\mathrm{time}}$ on the hidden-state transitions produced by the token-trained sequence model.

Because the backbone is not updated during the second phase, temporal gradients cannot modify its representations. The temporal-prediction head can therefore exploit only the temporal information already present in the token-trained hidden-state transitions. This regime provides a controlled baseline for isolating the effect of allowing temporal supervision to shape the backbone.
\subsection{Inference}

Generation proceeds autoregressively. Given the token history $\mathcal{X}_k$, the token head first predicts the next token $x_{k+1}$. After this token is processed by the backbone, the model obtains $h_{k+1}$ and computes $\Delta h_{k+1}$. The temporal head then predicts the associated inter-arrival time $\widehat{\Delta t}_{k+1}$ from $\Delta h_{k+1}$, and the timestamp is updated as $t_{k+1}=t_k+\widehat{\Delta t}_{k+1}$.

The token-decoding strategy and the point estimate used for temporal prediction are specified in \cref{sec:experimental-setup}.

\section{Experimental Setup}
\label{sec:experimental-setup}

Our experimental setup is designed to isolate the effect of allowing temporal supervision to update the shared backbone. Our primary evaluation therefore compares \joint{} and \twostage{} training, which use the same model components, data, temporal objective, and generation procedure within each domain, but differ in whether the temporal loss updates the shared backbone. This controlled comparison lets us determine whether this choice makes inter-arrival information more recoverable from the learned representations (\textbf{RQ1} in \cref{sec:recoverability}) and whether any such improvement preserves token-generation quality (\textbf{RQ2} in \cref{sec:results}).

We conduct this controlled comparison on four domains exhibiting distinct sequence structures, vocabularies, timestamp densities, and temporal regimes:
business-process traces, clinical event sequences, network traffic, and temporal knowledge graphs. Together, these four domains provide complementary tests of \textbf{RQ1} and \textbf{RQ2} across dense and partial timestamp coverage, short and long sequences, and substantially different temporal distributions. Appendix~\cref{app:domain-comparison} provides a detailed comparison of the structural and temporal properties of the four main evaluation domains, and Appendix~\cref{app:music-experiments} reports additional experiments on musical event sequences.

\subsection{Experimental Instantiation}
\label{ss:experimental-instantiation}

Unless stated otherwise, all experiments in the main paper instantiate \modelname{} with a Mamba2 backbone~\cite{dao_transformers_2024}. To assess the robustness of the controlled comparison to the choice of causal autoregressive backbone, and to illustrate that the framework can be instantiated beyond SSMs, Appendix~\cref{app:backbone-robustness} repeats the comparison with a GPT-2 backbone. The temporal-head architecture is selected separately for each domain to account for differences in sequence structure and temporal
distribution.

Complete domain-specific preprocessing settings, model architectures, temporal-head configurations, loss weights, and training hyperparameters are reported in Appendix~\cref{app:model-configurations}. Appendix~\cref{app:compute} reports the corresponding training-time and peak-memory overhead. Appendix~\cref{app:hyperparameter-sensitivity} reports an auxiliary-loss-weight sensitivity study for network traffic and temporal knowledge graph.

\subsection{Densely Timestamped Datasets}
\label{ss:densely-timestamped}

In this subsection, each sequence in $\mathcal{D}$ follows the canonical formulation introduced in \cref{ss:problem_formulation}: every token $x_k$ is associated with an observed timestamp $t_k$, so inter-arrival times are defined for every consecutive pair of events.

\parab{Business-process traces.}
\label{ss:bpi}
BPI Challenge 2018 \cite{vandongen2018bpi,mannhardt2018bpi} (BPI2018) is a business-process event log in which each case is represented as a sequence of timestamped activities. Each complete case forms one process trace. For every event, the activity label defines the token $x_k$, while its occurrence time defines the associated timestamp $t_k$.

Events are ordered chronologically within each trace. When multiple activities share the same timestamp, they are retained as separate tokens and therefore produce valid zero-valued inter-arrival targets. The original event log also contains resource identifiers, lifecycle transitions, application metadata, permit information, and other event- and case-level attributes, but we discard these auxiliary attributes and retain only the ordered activity labels and their timestamps to keep the modeling simple.

\parab{Clinical event sequences.}
\label{ss:mimic}
MIMIC-IV~\cite{johnson2023mimic} is a large-scale electronic health record dataset containing clinical observations collected during hospital and intensive-care stays. We use the \texttt{chartevents} table, which records timestamped bedside observations such as vital signs, ventilator settings, nursing assessments, and other charted clinical measurements. The event defines the token $x_k$, and the recording time defines the timestamp $t_k$.

Events are ordered chronologically within each stay. When multiple observations share the same timestamp, they are retained as separate tokens, yielding valid zero-valued inter-arrival times, as in BPI2018. We discard the remaining clinical and administrative attributes and retain only the ordered event identities and their timestamps.

\parab{Log-transformed temporal targets.}
\label{ss:dense-log-targets}
Although the temporal regimes differ across the two dense domains, both datasets exhibit highly heterogeneous inter-arrival times: BPI2018 contains very long delays, whereas MIMIC-IV is bursty and strongly zero-inflated, as summarized in Appendix~\cref{app:domain-comparison}. We therefore predict a log-transformed auxiliary target in both settings, preserving the monotone ordering of inter-arrival times while compressing large gaps so that they do not dominate the temporal loss and short gaps remain well resolved.

Because both datasets contain simultaneous events, the temporal prediction mechanism must accommodate observed inter-arrival times that are exactly zero. In particular, a plain $\log(\Delta t_k)$ would be undefined when
$\Delta t_k=0$. We therefore define
\begin{equation}
  z_k
  =
  \log\left(1+\frac{\Delta t_k}{s}\right),
  \label{eq:log-time-target}
\end{equation}
where $s$ is a dataset-specific time scale. This transformed quantity $z_k$ defines the auxiliary temporal target used for supervision. The temporal head produces a raw scalar output, which is mapped through a softplus nonlinearity to obtain a non-negative inter-arrival prediction $\widehat{\Delta t}_k$. This prediction is then transformed in the same way:
\begin{equation}
  \widehat{z}_k
  =
  \log\left(1+\frac{\widehat{\Delta t}_k}{s}\right).
  \label{eq:log-time-prediction}
\end{equation}

We instantiate the temporal factor as the
fixed-variance log-normal density
\begin{equation}
  f^{\mathrm{time}}_{\phi}
  \left(
    \Delta t_k; \Delta h_k
  \right)
  =
  \frac{1}{(s+\Delta t_k)\sqrt{2\pi\sigma^2}}
  \exp\left(
    -\frac{(z_k-\widehat{z}_k)^2}{2\sigma^2}
  \right),
  \label{eq:main-time-likelihood}
\end{equation}
where $\sigma^2$ is fixed. Its negative log-likelihood is equivalent, up to terms independent of the model parameters, to squared error between $\widehat{z}_k$ and $z_k$. This is a modeling choice rather than a requirement of the framework. Appendix~\cref{app:music-experiments} reports an auxiliary GiantMIDI experiment that instead applies squared-error regression to scaled raw inter-arrival times, testing whether the results depend on the logarithmic target transformation.

\subsection{Partially Timestamped Datasets}
\label{ss:partially-timestamped}

Many temporally structured sequences do not provide a timestamp for every token but only for selected structural boundaries. These settings are therefore important for evaluating whether temporal supervision can still shape useful representations when it is available only sparsely within the sequence.

In this subsection, we consider domains in which timestamps are attached only to selected boundary tokens rather than to every token in the sequence. Let $\mathcal{I}_{\mathrm{time}}$ denote the set of positions carrying valid temporal targets. For each $k\in\mathcal{I}_{\mathrm{time}}$, let $k^{-}$ denote the preceding position in $\mathcal{I}_{\mathrm{time}}$. The temporal target and its corresponding hidden-state representation are then defined as
\begin{equation}
  \Delta t_k=t_k-t_{k^{-}},
  \qquad
  \Delta h_k=h_k-h_{k^{-}}.
  \label{eq:partial-time-transitions}
\end{equation}

The temporal loss is evaluated only for $k\in\mathcal{I}_{\mathrm{time}}$. Tokens outside $\mathcal{I}_{\mathrm{time}}$ continue to contribute to the token-generation objective but do not contribute directly to the temporal objective. This defines a different setting from the canonical formulation introduced in
\cref{ss:problem_formulation}.

\parab{Network traffic.}
\label{ss:network-traffic}
Network traffic consists of timestamped packets exchanged between hosts and recorded in packet-capture files. We use captures of video-streaming traffic~\cite{networkdataset}, following the preprocessing procedure of Chu et al.~\cite{chu2025netssm}. Packet contents are serialized as sequences of byte tokens, with a dedicated \texttt{<|pkt|>} token delimiting successive packets. Timestamps are attached only to these \texttt{<|pkt|>} boundary tokens, so $\mathcal{I}_{\mathrm{time}}$ contains the packet-boundary positions.

\parab{Temporal knowledge graphs.}
\label{ss:dkg}
Temporal knowledge graphs represent evolving relational information as timestamped facts $(h,r,o,\tau)$, where $h$ and $o$ denote entities, $r$ denotes a relation, and $\tau$ denotes the fact timestamp. We use the GDELT temporal knowledge graph~\cite{renet}, which contains world events extracted from news media and organized chronologically.

Facts are mapped to a fixed vocabulary of entity and relation identifiers and represented as structured token blocks of the form \texttt{<|STM|>, h, r, o}. Consecutive facts are grouped into fixed-length training sequences. The timestamp of each fact is attached only to its \texttt{<|STM|>} marker, so $\mathcal{I}_{\mathrm{time}}$ contains the
\texttt{<|STM|>} positions.

\parab{Log-transformed temporal targets.}
To keep the temporal objective comparable across all domains, we apply the same target scaling, logarithmic transformation, softplus-constrained prediction, and squared-error loss introduced in \cref{ss:dense-log-targets}. For network traffic only, packet inter-arrival times are clipped at the 99th percentile before scaling. This reduces the influence of rare extreme delays on the dataset-specific scaling and training stability, beyond the compression already provided by the logarithmic transformation.

\subsection{Generation Settings}
\label{ss:generation-settings}

Token decoding is domain-specific: BPI2018, MIMIC-IV, and dynamic knowledge graphs use greedy decoding, whereas network traffic uses stochastic sampling, following Chu et al.~\cite{chu2025netssm}. Complete decoding parameters are reported in Appendix~\cref{app:model-configurations}. For temporal prediction, the raw output of the temporal head is passed through a softplus nonlinearity to obtain the non-negative point estimate $\widehat{\Delta t}_k$, which is added to the preceding generated timestamp.

\subsection{External Reference Baselines}
\label{ss:domain-specific-baselines}

For additional descriptive context, Appendix~\cref{app:baseline-details} reports one retrained domain-adapted baseline per dataset, including its architecture, probing adaptation, results, and interpretation. These external models are not used to answer \textbf{RQ1} or \textbf{RQ2}, because they differ from \modelname{} in architecture, objective, temporal representation, and generation procedure.

\section{Temporal Recoverability} \label{sec:recoverability}

This section addresses \textbf{RQ1 (Temporal Recoverability)}: whether \joint{} training makes inter-arrival information more recoverable from the sequence-backbone representations than \twostage{} training. To answer this question, we evaluate after training the same representation that is supplied to the temporal head. The backbone is frozen, and temporal information is extracted either with a linear probe or through a non-parametric analysis of representation geometry. No probe gradients are propagated into the backbone, so the resulting measurements reflect information already present in the learned representations.

\subsection{Recoverability Metrics} \label{ss:recoverability-metrics}

In order to respond to \textbf{RQ1}, we define \emph{temporal recoverability} as the extent to which the temporal target associated with an event transition can be extracted from frozen backbone representations using a simple readout or reflected in their geometry. For each valid temporal transition, we evaluate the representation supplied to the temporal head and the transformed inter-arrival target used during training.

\parab{Linear timing probe.}
The first recoverability metric consists of training a ridge-regression probe to predict the temporal target $z_{k}$ (see \cref{eq:log-time-target}) from the corresponding frozen temporal representation $r_{k}$ (see \cref{eq:hidden-transition}):
\begin{equation}
r_k \longrightarrow z_k.
\label{eq:linear-time-probe}
\end{equation}
Although this may resemble the second stage of \twostage{} training, the two procedures serve different purposes. The \twostage{} temporal head is part of the model itself and is optimized for downstream temporal prediction, whereas the probe is a post hoc diagnostic trained only after model training has finished. It uses a simple ridge-regression readout, does not reuse the trained temporal head, and is introduced solely to measure how easily temporal information can be extracted from the frozen representation.

We report mean absolute error (MAE) and the coefficient of determination (\(R^2\)). A lower MAE indicates that the temporal target can be reconstructed more accurately from the representation, while a higher \(R^2\) indicates that the probe explains a larger fraction of its variability. Lower MAE and higher \(R^2\) therefore indicate stronger temporal recoverability.

\parab{Temporal Cohesion Score.}
The second recoverability metric evaluates whether transitions associated with similar temporal targets occupy nearby regions of representation space. For each temporal representation \(r_k\), we identify its \(K\) nearest neighbors under cosine similarity and compute the mean absolute difference between their associated temporal targets:
\begin{equation}
d_{\mathrm{nn}} = \frac{1}{N K} \sum_{k=1}^{N}
\sum_{j\in\mathcal{N}_K(k)} |z_k-z_j|,
\end{equation}
where \(\mathcal{N}_K(k)\) denotes the \(K\) nearest neighbors of \(r_k\).

We compare this quantity with the mean target difference obtained by assigning \(K\) randomly selected transitions to each \(r_k\), denoted \(d_{\mathrm{rand}}\), and report
\begin{equation}
\mathrm{TCS} =
\frac{d_{\mathrm{rand}}-d_{\mathrm{nn}}}
{d_{\mathrm{rand}}+\varepsilon},
\end{equation}
where \(\varepsilon\) prevents division by zero. In all experiments, we set \(K=5\) and use cosine similarity. Sensitivity to the neighborhood size and similarity measure is reported in Appendix~\cref{app:metric-sensitivity}.

A high TCS indicates that transitions with similar inter-arrival times are substantially closer in representation space than expected under random pairing. Temporal information is therefore more strongly reflected in the
geometry of the learned representations. Appendix~\cref{app:representation-analysis} provides a separate descriptive
layer-wise alignment analysis of the same representation geometry across domains.

\subsection{Response to RQ1}
\label{ss:rq1-response}

Table~\ref{tab:temporal-recoverability} reports transition-level temporal recoverability across the four evaluation domains. In every domain, \joint{} achieves higher TCS and probe \(R^2\), together with lower probe MAE, than \twostage{}. Thus, joint training improves both local temporal coherence and the linear accessibility of inter-arrival information, under both dense and partial timestamp supervision.

The diagnostic profile nevertheless differs by domain. On BPI2018, TCS increases only slightly, from \(0.794\) to \(0.803\), while the probe improvements are clearer: \(R^2\) increases from \(0.461\) to \(0.631\) and MAE decreases from \(3.342\) to \(2.775\). This pattern may suggest that joint training makes temporal information more linearly accessible without substantially changing local temporal neighborhoods. MIMIC-IV shows a similar pattern. Network traffic, by contrast, shows large improvements in all three diagnostics, including TCS from \(0.147\) to \(0.622\), probe \(R^2\) from \(0.190\) to \(0.394\), and MAE from \(0.382\) to \(0.308\). The concurrent gains may suggest changes in both local temporal coherence and linear accessibility. For the temporal knowledge graph, the changes are smaller across all three diagnostics than on the other domains: unlike BPI2018 and MIMIC-IV, where the probe metrics improve more clearly than TCS, and network traffic, where all three diagnostics improve substantially, its gains remain modest but consistently favor \joint{}.
\begin{tcolorbox}[
    colback=gray!10,
    colframe=black,
    arc=4mm,
    boxrule=1.5pt,
    left=5pt,
    right=5pt,
    top=5pt,
    bottom=5pt
]
\textbf{Takeaways.} The controlled results provide an affirmative answer to \textbf{RQ1}: compared with \twostage{}, \joint makes inter-arrival information more recoverable from the representations supplied to the temporal head.
\end{tcolorbox}

\begin{table}[t]
  \centering
  \small
  \begin{tabular}{llrrr}
    \toprule
    Dataset
      & Method
      & TCS $\uparrow$
      & Probe \(R^2\) $\uparrow$
      & Probe MAE $\downarrow$ \\
    \midrule

    \multirow{2}{*}{BPI2018}
      & \joint{}
      & \textbf{0.8030}
      & \textbf{0.631}
      & \textbf{2.775} \\
      & \twostage{}
      & 0.7940
      & 0.461
      & 3.342 \\
    \midrule

    \multirow{2}{*}{MIMIC-IV}
      & \joint{}
      & \textbf{0.8620}
      & \textbf{0.6580}
      & \textbf{0.5970} \\
      & \twostage{}
      & 0.8480
      & 0.5880
      & 0.6450 \\
    \midrule

    \multirow{2}{*}{Network traffic}
      & \joint{}
      & \textbf{0.6216}
      & \textbf{0.3943}
      & \textbf{0.3080} \\
      & \twostage{}
      & 0.1468
      & 0.1902
      & 0.3824 \\
    \midrule

    \multirow{2}{*}{Temporal KG}
      & \joint{}
      & \textbf{0.2027}
      & \textbf{0.4449}
      & \textbf{904.14} \\
      & \twostage{}
      & 0.1722
      & 0.4296
      & 955.69 \\
    \bottomrule
  \end{tabular}

  \caption{
    Transition-level temporal recoverability across domains. Boldface
    identifies the better result within the controlled
    \joint{}--\twostage{} comparison.
  }
  \label{tab:temporal-recoverability}
\end{table}

\section{Effect on Generation Quality}
\label{sec:results}

The previous section provided an affirmative answer to \textbf{RQ1}, showing that \joint{} training makes temporal information more recoverable from the representations supplied to the temporal head. However, this improvement would be of limited value if it were obtained at the expense of the model's primary generative capability. We therefore turn to \textbf{RQ2} and test whether the additional temporal supervision introduced by \joint{} training systematically degrades generation quality relative to \twostage{} training across the four evaluation domains.

\subsection{Preservation Metrics}

We evaluate generated samples against held-out data using domain-specific protocols taken from the literature.

\parab{Business-process traces.}
For BPI2018, we evaluate generated activity sequences using Damerau--Levenshtein similarity (DLS), following the suffix-prediction protocol commonly used in predictive process monitoring~\cite{tax2017predictive}. For each evaluated prefix, similarity between the generated and reference suffixes is computed as
\[
1-\frac{d_{\mathrm{DL}}(\hat{\mathbf{x}},\mathbf{x})}
{\max(|\hat{\mathbf{x}}|,|\mathbf{x}|)},
\]
where ($d_{\mathrm{DL}}$) denotes Damerau--Levenshtein distance. We report the mean and median over all evaluated suffix comparisons. Higher values indicate that the generated continuation more closely reproduces the activity ordering of the corresponding reference trace.

\parab{Clinical event sequences.}
For MIMIC-IV, we compare the marginal distribution and coverage of generated clinical event types against held-out sequences. We report Jensen--Shannon divergence and total variation distance between the generated and reference event-token distributions, with lower values indicating closer agreement. We also report recall over the 50 most frequent reference event types, defined as the fraction of these event types that appear at least once in the generated data. Higher recall indicates better coverage of frequent clinical events.

\parab{Network traffic.}
We compare marginal distributions of key packet-header fields between generated and real traffic, following prior work~\cite{jiang2024netdiffusion,chu2025netssm}. We report Jensen--Shannon divergence for source and destination IP addresses and TCP ports. Lower divergence indicates that generated packets more faithfully reproduce the distributional characteristics of real network traffic.

\parab{Temporal knowledge graphs.}
We use filtered link prediction following~\cite{renet}. Given a query prefix, the model ranks candidate objects while masking facts that appear elsewhere in the dataset. We report mean reciprocal rank, mean rank, and Hits@({1,3,10}). Higher mean reciprocal rank and Hits scores, and lower mean rank, indicate that the model better captures the relational structure required to complete a fact.

\begin{table*}[t]
  \centering
  \small
  \setlength{\tabcolsep}{5pt}
  \begin{tabular}{llrrrrr}
    \toprule
    Dataset
      & Method
      & \multicolumn{5}{c}{Content-generation quality} \\
    \midrule

    \multirow{3}{*}{BPI2018}
      &
      & DLS mean $\uparrow$
      & DLS median $\uparrow$
      &
      &
      & \\
    \cmidrule(lr){2-7}

      & \joint{}
      & \textbf{0.487}
      & \textbf{0.500}
      &
      &
      & \\

      & \twostage{}
      & \textbf{0.487}
      & \textbf{0.500}
      &
      &
      & \\

    \midrule

    \multirow{3}{*}{MIMIC-IV}
      &
      & Token JSD $\downarrow$
      & Token TV $\downarrow$
      & Recall@50 $\uparrow$
      &
      & \\
    \cmidrule(lr){2-7}

      & \joint{}
      & 0.314
      & 0.477
      & \textbf{0.780}
      &
      & \\

      & \twostage{}
      & \textbf{0.302}
      & \textbf{0.451}
      & \textbf{0.780}
      &
      & \\

    \midrule

    \multirow{3}{*}{Network traffic}
      &
      & JSD ip.src $\downarrow$
      & JSD ip.dst $\downarrow$
      & JSD tcp.src $\downarrow$
      & JSD tcp.dst $\downarrow$
      & \\
    \cmidrule(lr){2-7}

      & \joint{}
      & \textbf{0.0412}
      & \textbf{0.0311}
      & \textbf{0.0764}
      & \textbf{0.1120}
      & \\

      & \twostage{}
      & 0.0669
      & 0.0516
      & 0.1248
      & 0.2183
      & \\

    \midrule

    \multirow{3}{*}{Temporal KG}
      &
      & MRR $\uparrow$
      & MR $\downarrow$
      & Hits@1 $\uparrow$
      & Hits@3 $\uparrow$
      & Hits@10 $\uparrow$ \\
    \cmidrule(lr){2-7}

      & \joint{}
      & \textbf{0.1461}
      & \textbf{269.06}
      & \textbf{0.0899}
      & \textbf{0.1490}
      & \textbf{0.2527} \\

      & \twostage{}
      & 0.0576
      & 2155.60
      & 0.0325
      & 0.0557
      & 0.1017 \\

    \bottomrule
  \end{tabular}

  \caption{
    Content-generation quality across domains. Boldface identifies the better
    result within the controlled
    \joint{}--\twostage{} comparison.
  }
  \label{tab:generation-quality}
\end{table*}

\subsection{Response to RQ2}
\label{ss:results}
Table~\ref{tab:generation-quality} reports content-generation quality across the four domains. The controlled comparison provides a negative answer to \textbf{RQ2}: \joint{} training does not systematically degrade generative performance relative to \twostage{} training. Its effect is nevertheless domain-dependent, but provides no evidence of a systematic degradation in generative performance.

On BPI2018, \joint{} and \twostage{} obtain identical mean and median DLS. The additional temporal supervision therefore changes neither the generated activity suffixes nor their similarity to the reference traces in this setting. On MIMIC-IV, \twostage{} achieves slightly lower token JSD and total variation distance, while both regimes obtain the same recall over the 50 most frequent clinical event types. Thus, \joint{} introduces a small degradation in matching the marginal event-token distribution, but does not reduce coverage of frequent clinical events. Appendix~\cref{app:validation-losses} reports the complementary held-out validation-loss diagnostics: MIMIC-IV is also the only main domain in which \joint{} has a slightly higher token loss than \twostage{} (1.296 versus 1.293, a difference of about 0.2\%). This small teacher-forced loss difference is consistent with the limited distributional change, but does not indicate a broad degradation in generation quality: frequent-event coverage is unchanged.

In contrast, \joint{} improves content-generation quality in both partially timestamped domains. On network traffic, \joint{} consistently lowers the JSD of all four evaluated packet-header fields, with particularly large improvements for source and destination TCP ports. On temporal knowledge graphs, \joint{} also improves every filtered link-prediction metric, yielding substantially higher MRR and Hits@\(\{1,3,10\}\), together with a markedly lower mean rank. In these domains, temporal supervision not only preserves content generation but appears to improve the structural regularities captured by the token generator. The lower held-out $L_{\mathrm{tok}}$ under \joint{} in both domains is consistent with these output-level improvements (see \Cref{app:validation-losses}).

\begin{tcolorbox}[
    colback=gray!10,
    colframe=black,
    arc=4mm,
    boxrule=1.5pt,
    left=5pt,
    right=5pt,
    top=5pt,
    bottom=5pt
]
\textbf{Takeaways.} The results provide a negative answer to \textbf{RQ2}.
Although \joint{} does not dominate \twostage{} on every individual metric, there is no evidence that the improved temporal recoverability established in \textbf{RQ1} is obtained at the cost of a systematic degradation in the model's primary generative capability.
\end{tcolorbox}

\section{Related Work}\label{sec:related}
The prediction problem formalized in \cref{eq:joint-factorization} has long been studied in the point-process literature~\cite{daley2008introduction}. Recent neural approaches have used recurrent architectures~\cite{boyd2020user,du2016recurrent} and transformers~\cite{panos2024decomposable} to model continuous-time event sequences. Our focus is on the stricter generative setting, where both event identity and timing must be produced autoregressively. Prior work in this domain typically either decouples timing from content generation or does not model timing at all. For example, some methods assign inter-arrival times with separate components after content generation~\cite{chu2025netssm,jiang2024netdiffusion}, whereas others focus on event generation without modeling inter-arrival times within the same autoregressive generator~\cite{yin2022practical,renet}. By contrast, \modelname{} focuses on how the event and temporal factors in \cref{eq:joint-factorization} interact when they are learned within the same autoregressive sequence model.

A key distinction is that, even when prior work models event identity and timing jointly within a common model~\cite{boyd2020user,du2016recurrent,panos2024decomposable}, it does not study how the temporal term \(p_{\theta,\phi}(\Delta t_{k+1}\mid x_{k+1}, \mathcal{X}_k)\) should interact with the event term \(p_{\theta,\alpha}(x_{k+1}\mid \mathcal{X}_k)\) from a representation-learning perspective when both are learned within the same model. This is the specific question studied in this paper.

\section{Discussion and Future Work}\label{sec:future}

We introduced \modelname, an autoregressive model that jointly predicts tokens and inter-event times from a shared backbone. Across four temporally structured domains, \joint{} training yields representations from which temporal information is more easily recoverable (\textbf{RQ1}). At the same time, it does not systematically degrade downstream generation quality relative to \twostage{} training (\textbf{RQ2}): generation quality is unchanged on BPI2018, modestly reduced on selected MIMIC-IV distributional metrics, and improved on network traffic and temporal knowledge graphs. These results indicate that allowing timing gradients to shape the shared backbone can improve temporal recoverability without imposing a systematic cost on generative performance.

\subsection{Linking Temporal Recoverability and Generation Quality}

The relationship between temporal recoverability and content-generation quality is domain-dependent. On BPI2018, improved temporal recoverability is accompanied by unchanged activity-sequence quality. On MIMIC-IV, it coexists with a modest degradation in marginal event-token distribution metrics, but no reduction in frequent-event coverage. On network traffic and temporal knowledge graphs, \joint{} training is associated with better evaluated content-generation metrics alongside improved temporal recoverability. These results indicate that stronger temporal organization does not necessarily impose a uniform temporal--semantic trade-off. Instead, its effect depends on the domain structure, the density of temporal supervision, and the content properties captured by the evaluation protocol.

\noindent \textbf{Validation loss analysis:} In addition to the analysis described above, we also carry out a direct check of the effectiveness of the temporal head in predicting inter-arrival times, given the true difference in token representations (see \cref{eq:time-loss}). We compute the temporal loss on the held-out validation data for each dataset (which has the true inter-arrival times), and report it in Appendix \cref{app:validation-losses}. We observe that the timing loss for most cases, is lower using the \joint{} method. However, this is a only a coarse measurement how well the time head has learned the underlying distribution of temporal data. A more careful check would require multiple, long generation sequences, appropriately calibrated against the training data, after which the temporal distributions would be compared for real and generated data. As we note in \cref{ss:future-work}, a rigorous evaluation pipeline for this would be an interesting direction for future work.

\subsection{Towards a Mechanistic Understanding}
\label{ss:mechanistic_understanding}
Appendix~\cref{app:representation-analysis} compares the layer-wise representations learned by the token-generation model with and without explicit temporal supervision. We use Centered Kernel Alignment (CKA), Centered Kernel Nearest Neighbor Alignment (CKNNA), and Mutual $k$-Nearest Neighbor (M-KNN). Across domains, temporal supervision changes the learned representations, but the form of this change varies: in some cases it affects both global and local geometry, whereas in others it is primarily local. Changes are also generally less pronounced in earlier layers, which may reflect their stronger focus on the input structure. The patterns nevertheless differ across the four domains, so they do not yet provide a single mechanistic account of how temporal supervision reshapes representation learning in SSMs.

\subsection{Future Work}
\label{ss:future-work}

Several questions remain about the conditions under which temporal supervision is beneficial. Building on the sensitivity analyses in Appendix~\cref{app:hyperparameter-sensitivity}, future work should test more systematically how the gains from \joint{} training depend on temporal-supervision strength across all domains, temporal-head capacity, and the structure of the target domain. This would clarify the conditions under which temporal gradients beneficially reshape the backbone.

The practical value of improved temporal recoverability also warrants direct evaluation. Future work should test whether it translates into better performance on downstream tasks that explicitly depend on temporal structure, such as anomaly detection, forecasting, or temporal retrieval. This would help establish whether the representation-level gains observed here are not only diagnostic but also practically useful across temporally structured domains.

Finally, future work should assess the distributional fidelity of timestamps produced during free autoregressive generation, testing whether the representation-level gains translate into faithful temporal synthesis beyond the content-generation metrics considered in this study. For \textbf{RQ2}, we prioritized domain-specific generation protocols established in the respective application literatures, allowing a controlled comparison of the two training regimes using metrics with existing task-specific meaning. These protocols primarily assess event content, however. Developing standardized evaluation procedures for jointly generated event content and timestamps is therefore an important next step, particularly because existing generative approaches often decouple temporal prediction from content generation (\cref{sec:related}).

\bibliographystyle{ACM-Reference-Format}
\bibliography{references}

@inproceedings{dao_transformers_2024,
author = {Dao, Tri and Gu, Albert},
title = {Transformers are SSMs: generalized models and efficient algorithms through structured state space duality},
year = {2024},
publisher = {JMLR.org},
booktitle = {Proceedings of the 41st International Conference on Machine Learning},
articleno = {399},
numpages = {31},
location = {Vienna, Austria},
series = {ICML'24}
}

@InProceedings{pmlr-v97-kornblith19a,
  title = 	 {Similarity of Neural Network Representations Revisited},
  author =       {Kornblith, Simon and Norouzi, Mohammad and Lee, Honglak and Hinton, Geoffrey},
  booktitle = 	 {Proceedings of the 36th International Conference on Machine Learning},
  pages = 	 {3519--3529},
  year = 	 {2019},
  editor = 	 {Chaudhuri, Kamalika and Salakhutdinov, Ruslan},
  volume = 	 {97},
  series = 	 {Proceedings of Machine Learning Research},
  month = 	 {09--15 Jun},
  publisher =    {PMLR},
  url = 	 {https://proceedings.mlr.press/v97/kornblith19a.html}
}

@inproceedings{NIPS2007_d5cfead9,
 author = {Gretton, Arthur and Fukumizu, Kenji and Teo, Choon and Song, Le and Sch\"{o}lkopf, Bernhard and Smola, Alex},
 booktitle = {Advances in Neural Information Processing Systems},
 editor = {J. Platt and D. Koller and Y. Singer and S. Roweis},
 pages = {},
 publisher = {Curran Associates, Inc.},
 title = {A Kernel Statistical Test of Independence},
 url = {https://proceedings.neurips.cc/paper_files/paper/2007/file/d5cfead94f5350c12c322b5b664544c1-Paper.pdf},
 volume = {20},
 year = {2007}
}

@InProceedings{pmlr-v235-huh24a,
  title = 	 {Position: The Platonic Representation Hypothesis},
  author =       {Huh, Minyoung and Cheung, Brian and Wang, Tongzhou and Isola, Phillip},
  booktitle = 	 {Proceedings of the 41st International Conference on Machine Learning},
  pages = 	 {20617--20642},
  year = 	 {2024},
  editor = 	 {Salakhutdinov, Ruslan and Kolter, Zico and Heller, Katherine and Weller, Adrian and Oliver, Nuria and Scarlett, Jonathan and Berkenkamp, Felix},
  volume = 	 {235},
  series = 	 {Proceedings of Machine Learning Research},
  month = 	 {21--27 Jul},
  publisher =    {PMLR},
  url = 	 {https://proceedings.mlr.press/v235/huh24a.html}
}

@article{ssm-better,
  title={The End of Transformers? On Challenging Attention and the Rise of Sub-Quadratic Architectures},
  author={Alexander Fichtl and Jeremias Bohn and Josefin Kelber and Edoardo Mosca and Georg Groh},
  journal={ArXiv},
  year={2025},
  volume={abs/2510.05364},
  url={https://api.semanticscholar.org/CorpusID:281886741}
}

@misc{ssm-paper,
      title={Efficiently Modeling Long Sequences with Structured State Spaces}, 
      author={Albert Gu and Karan Goel and Christopher Ré},
      year={2022},
      eprint={2111.00396},
      archivePrefix={arXiv},
      primaryClass={cs.LG},
      url={https://arxiv.org/abs/2111.00396}, 
}

@article{chu2025netssm,
author = {Chu, Andrew and Jiang, Xi and Liu, Shinan and Bhagoji, Arjun and Bronzino, Francesco and Schmitt, Paul and Feamster, Nick},
title = {NetSSM: Multi-Flow and State-Aware Network Trace Generation using State-Space Models},
year = {2026},
issue_date = {March 2026},
publisher = {Association for Computing Machinery},
address = {New York, NY, USA},
volume = {4},
number = {CoNEXT1},
url = {https://doi.org/10.1145/3786289},
doi = {10.1145/3786289},
journal = {Proc. ACM Netw.},
month = mar,
articleno = {6},
numpages = {24}
}

@article{flowchronicle,
author = {C\"{u}ppers, Joscha and Schoen, Adrien and Blanc, Gregory and Gimenez, Pierre-Francois},
title = {FlowChronicle: Synthetic Network Flow Generation through Pattern Set Mining},
year = {2024},
issue_date = {December 2024},
publisher = {Association for Computing Machinery},
address = {New York, NY, USA},
volume = {2},
number = {CoNEXT4},
url = {https://doi.org/10.1145/3696407},
doi = {10.1145/3696407},
journal = {Proc. ACM Netw.},
month = nov,
articleno = {26},
numpages = {20}
}

@misc{vandongen2018bpi,
  doi = {10.4121/UUID:3301445F-95E8-4FF0-98A4-901F1F204972},
  url = {https://data.4tu.nl/articles/_/12688355/1},
  author = {van Dongen, Boudewijn and Borchert, F. (Florian)},
  language = {en},
  title = {BPI Challenge 2018},
  publisher = {Eindhoven University of Technology},
  year = {2018},
  copyright = {Creative Commons Zero v1.0 Universal}
}

@inproceedings{mannhardt2018bpi,
  author = {Mannhardt, Felix and de Leoni, Massimiliano and Reijers, Hajo A. and van der Aalst, Wil M. P.},
  title = {The {BPI} Challenge 2018},
  booktitle = {Proceedings of the Business Process Intelligence Challenge 2018},
  year = {2018}
}

@article{johnson2023mimic,
  author = {Johnson, Alistair E. W. and Bulgarelli, Lucas and Pollard, Tom J. and Horng, Steven and Celi, Leo Anthony and Mark, Roger G.},
  title = {{MIMIC-IV}, a Freely Accessible Electronic Health Record Dataset},
  journal = {Scientific Data},
  volume = {10},
  number = {1},
  pages = {1},
  year = {2023},
  doi = {10.1038/s41597-022-01899-x}
}

@article{kong2022giantmidi,
  author = {Kong, Qiuqiang and Li, Bochen and Song, Xuchen and Meng, Qishuo and Barlow, Mark and Dixon, Simon},
  title = {{GiantMIDI-Piano}: A Large-Scale {MIDI} Dataset for Classical Piano Music},
  journal = {Transactions of the International Society for Music Information Retrieval},
  volume = {5},
  number = {1},
  pages = {87--102},
  year = {2022},
  doi = {10.5334/tismir.80}
}

@inproceedings{hawthorne2019maestro,
  author = {Hawthorne, Curtis and Stasyuk, Andriy and Roberts, Adam and Simon, Ian and Huang, Cheng-Zhi Anna and Dieleman, Sander and Elsen, Erich and Engel, Jesse and Eck, Douglas},
  title = {Enabling Factorized Piano Music Modeling and Generation with the {MAESTRO} Dataset},
  booktitle = {International Conference on Learning Representations},
  year = {2019},
  url={https://openreview.net/forum?id=r1lYRjC9F7}
}

@inproceedings{huang2019musictransformer,
  author = {Huang, Cheng-Zhi Anna and Vaswani, Ashish and Uszkoreit, Jakob and Shazeer, Noam and Simon, Ian and Hawthorne, Curtis and Dai, Andrew M. and Hoffman, Matthew D. and Dinculescu, Monica and Eck, Douglas},
  title = {Music Transformer: Generating Music with Long-Term Structure},
  booktitle = {International Conference on Learning Representations},
  year = {2019},
  url={https://openreview.net/forum?id=rJe4ShAcF7}
}

@inproceedings{lanvin,
author = {Lanvin, Maxime and Gimenez, Pierre-Fran\c{c}ois and Han, Yufei and Majorczyk, Fr\'{e}d\'{e}ric and M\'{e}, Ludovic and Totel, Eric},
title = {Towards Understanding Alerts raised by Unsupervised Network Intrusion Detection Systems},
year = {2023},
isbn = {9798400707650},
publisher = {Association for Computing Machinery},
address = {New York, NY, USA},
url = {https://doi.org/10.1145/3607199.3607247},
doi = {10.1145/3607199.3607247},
booktitle = {Proceedings of the 26th International Symposium on Research in Attacks, Intrusions and Defenses},
pages = {135–150},
numpages = {16},
location = {Hong Kong, China},
series = {RAID '23}
}

@inproceedings{fraud_detection,
author = {Paparrizos, John and Boniol, Paul and Liu, Qinghua and Palpanas, Themis},
title = {Advances in Time-Series Anomaly Detection: Algorithms, Benchmarks, and Evaluation Measures},
year = {2025},
isbn = {9798400714542},
publisher = {Association for Computing Machinery},
address = {New York, NY, USA},
url = {https://doi.org/10.1145/3711896.3736565},
doi = {10.1145/3711896.3736565},
booktitle = {Proceedings of the 31st ACM SIGKDD Conference on Knowledge Discovery and Data Mining V.2},
pages = {6151–6161},
numpages = {11},
location = {Toronto ON, Canada},
series = {KDD '25}
}

@inproceedings{renet,
    title = "Recurrent Event Network: Autoregressive Structure Inferenceover Temporal Knowledge Graphs",
    author = "Jin, Woojeong  and
      Qu, Meng  and
      Jin, Xisen  and
      Ren, Xiang",
    editor = "Webber, Bonnie  and
      Cohn, Trevor  and
      He, Yulan  and
      Liu, Yang",
    booktitle = "Proceedings of the 2020 Conference on Empirical Methods in Natural Language Processing (EMNLP)",
    month = nov,
    year = "2020",
    address = "Online",
    publisher = "Association for Computational Linguistics",
    url = "https://aclanthology.org/2020.emnlp-main.541/",
    doi = "10.18653/v1/2020.emnlp-main.541",
    pages = "6669--6683"
}

@book{goodfellow_dl_2016,
  title     = {Deep Learning},
  author    = {Goodfellow, Ian and Bengio, Yoshua and Courville, Aaron},
  year      = {2016},
  publisher = {MIT Press}
}

@article{networkdataset,
author = {Bronzino, Francesco and Schmitt, Paul and Ayoubi, Sara and Martins, Guilherme and Teixeira, Renata and Feamster, Nick},
title = {Inferring Streaming Video Quality from Encrypted Traffic: Practical Models and Deployment Experience},
year = {2019},
issue_date = {December 2019},
publisher = {Association for Computing Machinery},
address = {New York, NY, USA},
volume = {3},
number = {3},
url = {https://doi.org/10.1145/3366704},
doi = {10.1145/3366704},
journal = {Proc. ACM Meas. Anal. Comput. Syst.},
month = dec,
articleno = {56},
numpages = {25}
}

@article{jiang2024netdiffusion,
  title={Netdiffusion: Network data augmentation through protocol-constrained traffic generation},
  author={Jiang, Xi and Liu, Shinan and Gember-Jacobson, Aaron and Bhagoji, Arjun Nitin and Schmitt, Paul and Bronzino, Francesco and Feamster, Nick},
  journal={Proceedings of the ACM on Measurement and Analysis of Computing Systems},
  volume={8},
  number={1},
  pages={1--32},
  year={2024},
  publisher={ACM New York, NY, USA}
}

@inproceedings{yin2022practical,
  title={Practical gan-based synthetic ip header trace generation using netshare},
  author={Yin, Yucheng and Lin, Zinan and Jin, Minhao and Fanti, Giulia and Sekar, Vyas},
  booktitle={Proceedings of the ACM SIGCOMM 2022 Conference},
  pages={458--472},
  year={2022}
}

@book{daley2008introduction,
  title={An introduction to the theory of point processes: volume II: general theory and structure},
  author={Daley, Daryl J and Vere-Jones, David},
  year={2008},
  publisher={Springer}
}

@article{boyd2020user,
  title={User-dependent neural sequence models for continuous-time event data},
  author={Boyd, Alex and Bamler, Robert and Mandt, Stephan and Smyth, Padhraic},
  journal={Advances in Neural Information Processing Systems},
  volume={33},
  pages={21488--21499},
  year={2020}
  }

@article{panos2024decomposable,
  title={Decomposable transformer point processes},
  author={Panos, Aristeidis},
  journal={Advances in Neural Information Processing Systems},
  volume={37},
  pages={88932--88955},
  year={2024}
}

@inproceedings{du2016recurrent,
  title={Recurrent marked temporal point processes: Embedding event history to vector},
  author={Du, Nan and Dai, Hanjun and Trivedi, Rakshit and Upadhyay, Utkarsh and Gomez-Rodriguez, Manuel and Song, Le},
  booktitle={Proceedings of the 22nd ACM SIGKDD international conference on knowledge discovery and data mining},
  pages={1555--1564},
  year={2016}
}

@inproceedings{tax2017predictive,
  title={Predictive Business Process Monitoring with {LSTM} Neural Networks},
  author={Tax, Niek and Verenich, Ilya and La Rosa, Marcello and Dumas, Marlon},
  booktitle={Advanced Information Systems Engineering},
  series={Lecture Notes in Computer Science},
  volume={10253},
  pages={477--492},
  year={2017},
  publisher={Springer},
  doi={10.1007/978-3-319-59536-8_30}
}

@inproceedings{zuo2020thp,
  title={Transformer Hawkes Process},
  author={Zuo, Simiao and Jiang, Haoming and Li, Zichong and Zhao, Tuo and Zha, Hongyuan},
  booktitle={Proceedings of the 37th International Conference on Machine Learning},
  series={Proceedings of Machine Learning Research},
  volume={119},
  pages={11692--11702},
  year={2020},
  publisher={PMLR},
  url={https://proceedings.mlr.press/v119/zuo20a.html}
}

@misc{loshchilov2017decoupled,
      title={Decoupled Weight Decay Regularization}, 
      author={Ilya Loshchilov and Frank Hutter},
      year={2019},
      eprint={1711.05101},
      archivePrefix={arXiv},
      primaryClass={cs.LG},
      url={https://arxiv.org/abs/1711.05101}, 
}
\newpage
\appendix
\appendix

\AddToHook{env/table/begin}{\footnotesize}
\AddToHook{env/table*/begin}{\footnotesize}

\section*{Appendix Overview}

In this appendix, we provide additional experimental details, robustness analyses, and supporting results that complement the main paper. In particular, we (i) compare the experimental domains and document the model configurations, (ii) evaluate the robustness of the results across datasets, backbones, temporal-loss weights, and temporal representations, (iii) report held-out validation-loss diagnostics, (iv) provide comparisons with domain-specific baselines, and (v) further analyze the recoverability metrics and the layer-wise geometry of the learned representations. The appendix is organized as follows:

\begin{enumerate}
    \item Comparison of the experimental domains
    (\cref{app:domain-comparison}).
    \item Additional experiments on symbolic music
    (\cref{app:music-experiments}).
    \item Application to an alternative autoregressive backbone
    (\cref{app:backbone-robustness}).
    \item Held-out validation-loss diagnostics for the main experiments
    (\cref{app:validation-losses}).
    \item Sensitivity to the auxiliary temporal-loss weight
    (\cref{app:hyperparameter-sensitivity}).
    \item Comparison of alternative temporal representations
    (\cref{app:temporal-representations}).
    \item Computational overhead
    (\cref{app:compute}).
    \item Domain-specific model configurations
    (\cref{app:model-configurations}).
    \item Domain-specific baseline setup and results
    (\cref{app:baseline-details}).
    \item Sensitivity of TCS to its metric parameters
    (\cref{app:metric-sensitivity}).
    \item Layer-wise representation analysis
    (\cref{app:representation-analysis}).
\end{enumerate}

\section{Comparison of Experimental Domains}
\label{app:domain-comparison}

This appendix expands on the four-domain evaluation introduced in \cref{ss:densely-timestamped} and \cref{ss:partially-timestamped}, summarizing  why the evaluation spans genuinely different settings. The main paper compares two densely timestamped domains (BPI2018 and MIMIC-IV) and two partially timestamped domains (network traffic and temporal knowledge graphs), while the auxiliary GiantMIDI experiment adds a dense musical setting. These datasets differ not only in application domain, but also in the granularity of the atomic event, the meaning and size of the token vocabulary, the scale of the input sequences, the positions at which timestamps are available, and the characteristic shape of the temporal distribution.

\begin{table*}[t]
  \centering
  \adjustbox{max width=\textwidth}{%
  \begin{tabular}{p{2.7cm}p{2.6cm}p{2.5cm}p{2.3cm}p{2.5cm}p{2.6cm}}
    \toprule
      & \multicolumn{2}{c}{Partial Timestamping}
      & \multicolumn{3}{c}{Dense Timestamping} \\
    \cmidrule(lr){2-3} \cmidrule(lr){4-6}
    Property
      & Temporal KG
      & Network Traffic
      & BPI2018
      & MIMIC-IV
      & GiantMIDI (Aux.) \\
    \midrule
    Atomic event
      & Knowledge-graph fact
      & Packet
      & Business activity
      & Charted clinical event
      & Piano note \\
    Token semantics
      & Entity / relation
      & Byte value
      & Activity label
      & \texttt{itemid}
      & Note pitch \\
    Vocabulary
      & 15,626 symbolic tokens
      & 261 byte-level tokens
      & 41 activities
      & 500 retained \texttt{itemid}s
      & 84 note tokens \\
    Typical sequence length
      & Hundreds to thousands
      & Tens of thousands
      & Tens
      & Thousands
      & Thousands \\
    Timestamp availability
      & Fact boundaries only
      & Packet boundaries only
      & Every event
      & Every event
      & Every note \\
    Temporal supervision
      & Sparse
      & Sparse
      & Dense
      & Dense
      & Dense \\
    Typical inter-arrival scale
      & Seconds
      & Microseconds
      & Minutes to days
      & minutes, with many zeros
      & Milliseconds--seconds \\
    Temporal distribution
      & Moderate variability
      & Heavy-tailed
      & Strongly heavy-tailed
      & Zero-inflated and bursty
      & Dense with many zero \\
    Simultaneity
      & No
      & No
      & common
      & extremely common
      & very common \\
    Outlier handling
      & No clipping
      & 99th-percentile clipping
      & No clipping
      & No clipping
      & No clipping \\
    Main temporal challenge
      & Sparse supervision
      & Extreme density
      & Very long delays
      & Bursty zero-inflated timing
      & Simultaneous events \\
    \bottomrule
  \end{tabular}}
  \caption{High-level comparison of the main evaluation domains and the
  auxiliary GiantMIDI benchmark.}
  \label{tab:domain-comparison}
\end{table*}

In \Cref{tab:domain-comparison}, we see that the datasets differ strongly even within each timestamping regime. Among the dense domains, BPI2018 is comparatively short, template-like, and dominated by long waiting periods between process stages, whereas MIMIC-IV is much longer, more heterogeneous, and strongly zero-inflated because many clinical events share the same chart time. GiantMIDI is also dense, but it operates at a much finer temporal scale, with sub-second note timing and many simultaneous events due to chords. Among the partial domains, temporal knowledge graphs use a large symbolic vocabulary and relatively coarse fact-level timing, whereas network traffic uses a much smaller byte-level vocabulary and extremely dense packet timing.

These differences matter directly for the interpretation of the main results. Because the same qualitative effect of joint token--time training appears across these settings, it is less likely to be an artifact of a single temporal scale, single supervision pattern, or single type of Two-Stage structure.

\section{Music Event Experiments}
\label{app:music-experiments}

\parab{Dataset and event representation.}
We use GiantMIDI~\cite{kong2022giantmidi}, a large-scale corpus of classical piano MIDI performances. Each piece is converted into an ordered sequence of note events as in MAESTRO-style symbolic music modeling~\cite{hawthorne2019maestro}: the token $x_k$ is note pitch and the timestamp $t_k$ is note onset. We retain only pitch events and onset times and discard note duration, velocity, composer and performer identifiers, YouTube metadata, and transcription metadata. As noted in \cref{ss:dense-log-targets}, this auxiliary experiment also differs from the main dense-domain setup in its temporal supervision: instead of the log-gap target, it regresses directly on scaled raw inter-arrival times. More specifically, we define the scaled target
\begin{equation}
  u_k = \frac{\Delta t_k}{s_{\mathrm{music}}},
  \qquad
  \widehat{u}_k =
  \operatorname{softplus}\!\left(g_{\phi}(\Delta h_k)\right),
  \label{eq:music-time-target}
\end{equation}
and instantiate the temporal factor as
\begin{equation}
  f^{\mathrm{time}}_{\phi}
  \left(
    \Delta t_k; \Delta h_k
  \right)
  =
  \frac{1}{s_{\mathrm{music}}\sqrt{2\pi\sigma^2}}
  \exp\left(
    -\frac{(u_k-\widehat{u}_k)^2}{2\sigma^2}
  \right),
  \label{eq:music-time-likelihood}
\end{equation}
with fixed variance $\sigma^2$. Thus, its negative log-likelihood is equivalent, up to parameter-independent terms, to squared error on scaled raw inter-arrival times.

\parab{Generation-quality evaluation.}
We evaluate generated music against held-out MIDI sequences with a window-based overlapping-area (OA) protocol inspired by the evaluation used by \cite{huang2019musictransformer}. Each sequence is segmented into fixed 2-second windows, and for every window we compute four simple symbolic music statistics: note density, pitch range, mean pitch, and pitch variance. For each feature, we compare the empirical distribution over generated windows with the corresponding held-out distribution through their overlapping area, where higher values indicate closer agreement. We report both the feature-wise OA scores and their mean.

\begin{table*}[t]
  \centering
  \adjustbox{max width=\textwidth}{%
  \begin{tabular}{llrrrrrrrr}
    \toprule
    Dataset
      & Method
      & OA mean $\uparrow$
      & Mean pitch OA $\uparrow$
      & Note density OA $\uparrow$
      & Pitch range OA $\uparrow$
      & Pitch var. OA $\uparrow$
      & TCS $\uparrow$
      & Probe \(R^2\) $\uparrow$
      & Probe MAE $\downarrow$ \\
    \midrule
    \multirow{2}{*}{GiantMIDI}
      & \joint{}
      & \textbf{0.172}
      & \textbf{0.345}
      & \textbf{0.230}
      & \textbf{0.050}
      & \textbf{0.063}
      & \textbf{-0.044}
      & \textbf{0.160}
      & \textbf{248.006} \\
    & \twostage{}
      & 0.016
      & 0.000
      & 0.063
      & 0.000
      & 0.000
      & \textbf{-0.044}
      & 0.142
      & 261.097 \\
    \bottomrule
  \end{tabular}}
  \caption{
    Auxiliary GiantMIDI results. OA denotes the mean and feature-wise
    overlapping-area scores computed from 2-second window statistics. Boldface
    identifies the better result within the controlled
    \joint{}--\twostage{} comparison.
  }
  \label{tab:giantmidi-results}
\end{table*}

\parab{Response.}
\Cref{tab:giantmidi-results} provides the same controlled comparison for the auxiliary musical setting. From the perspective of \textbf{RQ1}, temporal recoverability remains weak for both \modelname{} variants: TCS is identical (\(-0.044\) for both), but the probe scores still slightly favor \joint{} (\(R^2=0.160\) vs.\ \(0.142\), MAE \(248.006\) vs.\ \(261.097\)). Thus, even in this dense symbolic-music setting, joint training yields a modest recoverability advantage under the probe-based diagnostics.

From the perspective of \textbf{RQ2}, the controlled comparison shows no degradation of free-generation quality under \joint{} in this auxiliary setting. At the aggregate level, mean OA increases from \(0.016\) under \twostage{} training to \(0.172\) under \joint{} training. The same pattern holds for all four reported symbolic-music features. In particular, \joint{} improves mean-pitch OA from \(0.000\) to \(0.345\), note-density OA from \(0.063\) to \(0.230\), pitch-range OA from \(0.000\) to \(0.050\), and pitch-variance OA from \(0.000\) to \(0.063\). This mirrors the pattern observed for network traffic and temporal knowledge graphs in \cref{sec:results}, where \joint{} training is also associated with better evaluated content-generation metrics relative to \twostage{}.

\section{Alternative Autoregressive Backbone}
\label{app:backbone-robustness}

As described in \cref{ss:problem_formulation}, \modelname{} is not tied to a State Space Model: its shared backbone can be any causal autoregressive architecture that produces token-wise hidden states. This appendix therefore illustrates the framework with a GPT-2 backbone. We repeat the controlled \joint{}--\twostage{} comparison while keeping the datasets and evaluation protocols unchanged. \Cref{tab:gpt2-backbone} summarizes the results, so we keep the discussion brief here.

\begin{table}[t]
  \centering
  \footnotesize
  \setlength{\tabcolsep}{2pt}
  \begin{tabular}{@{}p{0.4cm}p{1.2cm}p{2.3cm}rr@{}}
    \toprule
    & Dataset & Metric & \joint{} & \twostage{} \\
    \midrule
    \multirow{13}{*}{\rotatebox{90}{Generation}}
      & \multirow{2}{*}{BPI2018} & Mean DLS $\uparrow$ & 0.486951 & \textbf{0.487380} \\
      & & Med. DLS $\uparrow$ & \textbf{0.500} & \textbf{0.500} \\
    \cmidrule(lr){2-5}
      & \multirow{2}{*}{MIMIC-IV} & JSD $\downarrow$ & 0.296 & \textbf{0.286} \\
      & & R@50 $\uparrow$ & \textbf{0.820} & 0.780 \\
    \cmidrule(lr){2-5}
      & \multirow{4}{*}{Network} & JSD(srcIP) $\downarrow$ & 0.5882 & \textbf{0.3902} \\
      & & JSD(dstIP) $\downarrow$ & 0.6931 & \textbf{0.2342} \\
      & & JSD(sPort) $\downarrow$ & 0.6931 & \textbf{0.2408} \\
      & & JSD(dPort) $\downarrow$ & 0.6931 & \textbf{0.4259} \\
    \cmidrule(lr){2-5}
      & \multirow{5}{*}{TKG} & MRR $\uparrow$ & 0.1310 & \textbf{0.1329} \\
      & & MR $\downarrow$ & \textbf{510.87} & 534.30 \\
      & & Hits@1 $\uparrow$ & 0.0711 & \textbf{0.0723} \\
      & & Hits@3 $\uparrow$ & 0.1385 & \textbf{0.1394} \\
      & & Hits@10 $\uparrow$ & 0.2439 & \textbf{0.2492} \\
    \midrule
    \multirow{12}{*}{\rotatebox{90}{Recoverability}}
      & \multirow{3}{*}{BPI2018} & TCS $\uparrow$ & -0.163 & \textbf{0.282} \\
      & & Probe \(R^2\) $\uparrow$ & \textbf{0.661} & 0.603 \\
      & & Probe MAE $\downarrow$ & \textbf{2.630} & 2.806 \\
    \cmidrule(lr){2-5}
      & \multirow{3}{*}{MIMIC-IV} & TCS $\uparrow$ & \textbf{0.876} & 0.857 \\
      & & Probe \(R^2\) $\uparrow$ & \textbf{0.663} & 0.607 \\
      & & Probe MAE $\downarrow$ & \textbf{0.595} & 0.627 \\
    \cmidrule(lr){2-5}
      & \multirow{3}{*}{Network} & TCS $\uparrow$ & 0.6465 & \textbf{0.6951} \\
      & & Probe \(R^2\) $\uparrow$ & \textbf{0.4297} & 0.3665 \\
      & & Probe MAE $\downarrow$ & \textbf{0.0967} & 0.1188 \\
    \cmidrule(lr){2-5}
      & \multirow{3}{*}{TKG} & TCS $\uparrow$ & \textbf{0.5242} & 0.5156 \\
      & & Probe \(R^2\) $\uparrow$ & 0.6057 & \textbf{0.6129} \\
      & & Probe MAE $\downarrow$ & 839.75 & \textbf{833.01} \\
    \bottomrule
  \end{tabular}
  \caption{
    GPT-2 backbone instantiation under the controlled \joint{}--\twostage{}
    comparison. Boldface
    identifies the better result for each metric.
  }
  \label{tab:gpt2-backbone}
\end{table}

\parab{Response.}
The GPT-2 results are more heterogeneous than the Mamba2 results. On MIMIC-IV and network traffic, the recoverability picture is more favorable to \joint{} than the generation picture. On MIMIC-IV, \joint{} clearly improves the timing-recovery metrics, but the generation comparison is mixed: \twostage{} achieves slightly lower JSD, whereas \joint{} improves recall over the 50 most frequent event types. On network traffic, the tension is sharper: \joint{} improves the linear-probe recoverability metrics, but \twostage{} is clearly better on all packet-header distribution-matching metrics, and even TCS favors \twostage{}. On BPI2018 and the temporal knowledge graph, GPT-2 does not reproduce the Mamba2 recoverability pattern. On BPI2018, generation quality is effectively unchanged under DLS. TCS favors \twostage{}, whereas both probe diagnostics favor \joint{}. On the temporal knowledge graph, the two regimes are effectively tied overall, with \joint{} improving TCS and mean rank but \twostage{} remaining slightly better on the probe metrics and the other link-prediction scores.

Overall, the GPT-2 appendix does not support a uniform story. In some domains, \joint{} yields more recoverable temporal information only with mixed or weaker generation results; in others, it does not even provide a clear recoverability advantage. The favorable qualitative picture in the main paper is therefore not reproduced consistently with a GPT-2 backbone. One plausible explanation is architectural: Mamba2's selective recurrent state dynamics may provide a more suitable inductive bias for representing temporal transitions in long event sequences, whereas GPT-2 distributes sequence history through attention-based representations. This interpretation remains tentative. The present comparison does not isolate architectural effects from differences in model capacity, optimization, or backbone-specific hyperparameter choices, so it does not establish that one architecture is intrinsically simpler or better suited to temporal supervision than the other.

\section{Held-Out Validation-Loss Diagnostics}
\label{app:validation-losses}

Table~\ref{tab:validation-losses} reports validation losses on the predefined validation partitions used in the four main experiments. We use the dataset-provided training and validation splits; validation samples are excluded from gradient-based training. Both losses are evaluated under observed validation histories and therefore characterize conditional prediction rather than free-running generation. Consequently, the reported losses neither compare generated sequences or timestamps with their real-data distributions nor capture errors that accumulate when the model conditions on its own generated history. They provide complementary optimization diagnostics, but do not by themselves establish high-quality free-running content generation or distributionally faithful timestamp generation.

\begin{table}[t]
  \centering
  \setlength{\tabcolsep}{4pt}
  \begin{tabular}{llrr}
    \toprule
    Dataset & Method & $L_{\mathrm{tok}}\downarrow$ & $L_{\mathrm{time}}\downarrow$ \\
    \midrule
    \multirow{2}{*}{BPI2018}
      & \joint{} & \textbf{1.066} & \textbf{7.813} \\
      & \twostage{} & 1.107 & 19.585 \\
    \cmidrule(lr){1-4}
    \multirow{2}{*}{MIMIC-IV}
      & \joint{} & 1.296 & \textbf{0.348} \\
      & \twostage{} & \textbf{1.293} & 0.446 \\
    \cmidrule(lr){1-4}
    \multirow{2}{*}{Network traffic}
      & \joint{} & \textbf{1.087} & \textbf{0.001370} \\
      & \twostage{} & 1.145 & 0.001692 \\
    \cmidrule(lr){1-4}
    \multirow{2}{*}{Temporal KG}
      & \joint{} & \textbf{4.718} & 0.239 \\
      & \twostage{} & 5.868 & \textbf{0.078} \\
    \bottomrule
  \end{tabular}
  \caption{Held-out validation losses for the four main experiments.
  Boldface identifies the lower result within each dataset and loss.}
  \label{tab:validation-losses}
\end{table}

For token prediction, \joint{} obtains lower validation $L_{\mathrm{tok}}$ on BPI2018, network traffic, and temporal knowledge graphs, whereas \twostage{} is marginally lower on MIMIC-IV. This pattern provides complementary context for the content-generation results in \cref{sec:results}, but does not replace the free-generation evaluation.

For temporal point prediction, \joint{} obtains lower $L_{\mathrm{time}}$ on BPI2018, MIMIC-IV, and network traffic. On the temporal knowledge graph, however, \twostage{} achieves lower direct temporal-head loss. This result should be interpreted narrowly:
$L_{\mathrm{time}}$ measures the conditional point-prediction error of the particular trained temporal head, rather than a general measure of temporal generation quality.

\section{Effect of Auxiliary-Loss Weight}
\label{app:hyperparameter-sensitivity}

This section studies the effect of the auxiliary temporal-loss weight $\lambda_{\mathrm{\tau}}$ on the two partially timestamped domains: network traffic and temporal knowledge graphs. The goal is to assess how increasing the strength of temporal supervision changes temporal recoverability and downstream generation quality.

\begin{table}[t]
  \centering
  \footnotesize
  \setlength{\tabcolsep}{2pt}
  \begin{tabular}{lccccccc}
    \toprule
    $\lambda_{\mathrm{\tau}}$ & TCS$\uparrow$ & $R^2\uparrow$ & MAE$\downarrow$
    & MRR$\uparrow$ & MR$\downarrow$ & H@1$\uparrow$ & H@10$\uparrow$ \\
    \midrule
    0.05 & 0.188 & 0.265 & 965.9 & 0.095 & 684.6 & 0.059 & 0.160 \\
    0.10 & 0.186 & 0.299 & 942.0 & 0.095 & 498.5 & 0.054 & 0.169 \\
    0.20 & 0.195 & 0.298 & 922.8 & 0.119 & 360.1 & 0.068 & 0.212 \\
    0.35 & 0.190 & 0.297 & 934.8 & 0.126 & 316.9 & 0.071 & 0.230 \\
    0.50 & 0.214 & 0.309 & 921.8 & 0.135 & 274.2 & 0.079 & 0.240 \\
    0.75 & 0.203 & 0.303 & 928.1 & 0.149 & 273.5 & 0.093 & 0.254 \\
    1.00 & 0.224 & 0.312 & 913.1 & 0.148 & 265.6 & 0.090 & 0.256 \\
    \bottomrule
  \end{tabular}
  \caption{Effect of the auxiliary temporal-loss weight
  $\lambda_{\mathrm{\tau}}$ on temporal recoverability and link prediction
  for the temporal knowledge graph. Hits@3 is omited for space reasons.}
  \label{tab:lambda-aux-dkg}
\end{table}

\begin{table}[t]
  \centering
  \setlength{\tabcolsep}{2pt}
  \adjustbox{max width=\columnwidth}{%
    \begin{tabular}{lccccccc}
      \toprule
      $\lambda_{\mathrm{\tau}}$ & TCS$\uparrow$ & $R^2\uparrow$ & MAE$\downarrow$ & JSD(src)$\downarrow$ & JSD(dst)$\downarrow$ & JSD(sport)$\downarrow$ & JSD(dport)$\downarrow$ \\
      \midrule
      10 & 0.323 & 0.385 & 0.331 & 0.300 & 0.182 & 0.231 & 0.419 \\
      50 & 0.289 & 0.294 & 0.346 & 0.278 & 0.168 & 0.218 & 0.374 \\
      100 & 0.606 & 0.377 & 0.316 & 0.187 & 0.112 & 0.203 & 0.351 \\
      200 & 0.377 & 0.288 & 0.363 & 0.213 & 0.127 & 0.202 & 0.363 \\
      500 & 0.614 & 0.330 & 0.344 & 0.341 & 0.216 & 0.240 & 0.400 \\
      \bottomrule
    \end{tabular}%
  }
  \caption{Effect of the auxiliary temporal-loss weight
  $\lambda_{\mathrm{\tau}}$ on recoverability and packet-header distribution
  matching for network traffic. JSD(src), JSD(dst), JSD(sport), and
  JSD(dport) denote the source/destination IP and source/destination TCP-port
  marginals, respectively.}
  \label{tab:lambda-aux-network}
\end{table}

The two domains exhibit different sensitivity profiles. On the temporal knowledge graph, increasing $\lambda_{\mathrm{\tau}}$ produces a fairly consistent improvement in both recoverability and downstream link prediction. \Cref{tab:lambda-aux-dkg} shows both trends. The strongest setting, $\lambda_{\mathrm{\tau}}=1.00$, yields the best TCS, $R^2$, MAE, mean rank, and Hits@10, while $\lambda_{\mathrm{\tau}}=0.75$ slightly improves MRR and H@1.

On network traffic, the dependence is less monotone, and no single value of $\lambda_{\mathrm{\tau}}$ optimizes all recoverability diagnostics. The highest TCS is obtained at $\lambda_{\mathrm{\tau}}=500$, whereas the highest probe ($R^2$) occurs at $\lambda_{\mathrm{\tau}}=10$ and the lowest probe MAE at $\lambda_{\mathrm{\tau}}=100$. \Cref{tab:lambda-aux-network} further shows that $\lambda_{\mathrm{\tau}}=100$ yields the best packet-header distribution-matching metrics. These results therefore do not reveal a simple trade-off between temporal recoverability and generation quality; rather, the different recoverability diagnostics respond differently to the strength of temporal supervision, while an intermediate weight provides the best agreement with the evaluated packet-header marginals.

\parab{Rule of Thumb for Selecting the Auxiliary-Loss Weight.}
These results indicate that there is no universally optimal value of $\lambda_{\mathrm{\tau}}$. For the main experiments, we chose the auxiliary-loss weight so that the weighted temporal term, $\lambda_{\mathrm{\tau}}L_{\mathrm{time}}$, was approximately of the same order of magnitude as the token-prediction loss $L_{\mathrm{tok}}$ during training. This scale-matching criterion prevents either term from numerically dominating the joint objective in \cref{eq:joint-objective}.

\section{Alternative Temporal Representations}
\label{app:temporal-representations}

The default temporal representation \(r_k\) introduced in \cref{ss:model_design} (\cref{eq:hidden-transition}) is defined as the difference between the hidden states associated with the two endpoints of a temporally supervised transition. We compare it with the alternative representation \(r'_k\), formed by concatenating the same endpoint states:
\begin{equation}
  r_k = \Delta h_k = h_k - h_{k-1},
  \qquad
  r'_k = [h_{k-1}; h_k].
\end{equation}
The representation \(r_k\) has dimension \(d\), whereas \(r'_k\) has dimension \(2d\) and therefore changes the input projection of the temporal head. We evaluate both choices on one densely timestamped domain (BPI2018) and one partially timestamped domain (network traffic), retaining the same backbone, data splits, temporal-head family, and optimization schedule within each representation.

On network traffic, using \(r'_k\) empirically changes the magnitude and optimization behavior of the temporal objective. Stable training therefore requires \(\lambda_{\tau}=50\), rather than the value \(\lambda_{\tau}=300\) used with \(r_k\). Consequently, the network experiment should be interpreted as a comparison between two practical temporal-input configurations rather than as a strictly representation-only ablation. The results obtained with \(r'_k\) use transition-level recoverability diagnostics.

\begin{table*}[t]
  \centering
  \setlength{\tabcolsep}{3pt}
  \adjustbox{max width=\textwidth}{%
  \begin{tabular}{llccccccccc}
    \toprule
    Dataset & Method
      & \multicolumn{3}{c}{Temporal recoverability}
      & \multicolumn{2}{c}{BPI2018 generation}
      & \multicolumn{4}{c}{Network-traffic generation} \\
    \cmidrule(lr){3-5}\cmidrule(lr){6-7}\cmidrule(lr){8-11}
      &
      & TCS $\uparrow$ & Probe $R^2$ $\uparrow$ & Probe MAE $\downarrow$
      & DLS mean $\uparrow$ & DLS median $\uparrow$
      & JSD ip.src $\downarrow$ & JSD ip.dst $\downarrow$
      & JSD tcp.src $\downarrow$ & JSD tcp.dst $\downarrow$ \\
    \midrule
    \multirow{2}{*}{BPI2018}
      & \joint{} & 0.7924 (\textit{0.8030}) & 0.8753 (\textit{0.631}) & 4,919,732 (\textit{2.775})
      & \textbf{0.487} (\textit{0.487}) & \textbf{0.500} (\textit{0.500})
      & -- & -- & -- & -- \\
      & \twostage{} & \textbf{0.8029} (\textit{0.7940}) & \textbf{0.8789} (\textit{0.461}) & \textbf{4,861,086} (\textit{3.342})
      & \textbf{0.487} (\textit{0.487}) & \textbf{0.500} (\textit{0.500})
      & -- & -- & -- & -- \\
    \midrule
    \multirow{2}{*}{Network traffic}
      & \joint{} & \textbf{0.2987} (\textit{0.6216}) & \textbf{0.0313} (\textit{0.3943}) & \textbf{0.00610} (\textit{0.3080})
      & -- & --
      & \textbf{0.6931} (\textit{0.0412}) & \textbf{0.6931} (\textit{0.0311})
      & 0.3477 (\textit{0.0764}) & 0.4344 (\textit{0.1120}) \\
      & \twostage{} & 0.2538 (\textit{0.1468}) & 0.0116 (\textit{0.1902}) & 0.00654 (\textit{0.3824})
      & -- & --
      & \textbf{0.6931} (\textit{0.0669}) & \textbf{0.6931} (\textit{0.0516})
      & \textbf{0.3431} (\textit{0.1248}) & \textbf{0.3926} (\textit{0.2183}) \\
    \bottomrule
  \end{tabular}%
  }
  \caption{Results with concatenation
  $r'_k=[h_{k-1};h_k]$ as the temporal representation. Parenthesized italic values
  report the results from the main paper for reference.
  Boldface identifies the better concatenation result within each
  \joint{}--\twostage{} comparison; ``--'' marks metrics not applicable to
  that dataset.}
  \label{tab:temporal-representation-results}
\end{table*}

\Cref{tab:temporal-representation-results} shows that the configuration based on \(r'_k\) does not reproduce the empirical pattern obtained with the default representation \(r_k\).

From the perspective of \textbf{RQ1} (\cref{sec:recoverability}), BPI2018 remains effectively a tie. With \(r'_k\), \twostage{} is numerically better on all three recoverability diagnostics, but the differences are small. Using \(r'_k\) therefore removes, rather than meaningfully reverses, the clearer \joint{} advantage obtained with \(r_k\). Generation quality is unchanged, with identical mean and median DLS under both regimes.

The network-traffic results show a clearer departure from the default configuration. With \(r'_k\), \joint{} still outperforms \twostage{} on the recoverability diagnostics, but the separation is substantially smaller, especially for TCS and probe \(R^2\). The representation \(r'_k\) therefore fails to reproduce the strong recoverability advantage observed with \(r_k\).

From the perspective of \textbf{RQ2} (\cref{sec:results}), both variants using \(r'_k\) also match the packet-header distributions poorly. Source- and destination-IP JSD reach \(0.6931\) under both regimes, while the TCP-port divergences are substantially higher than with \(r_k\). Because this degradation occurs for both \joint{} and \twostage{}, it does not indicate a Joint-specific trade-off, but rather that the tested configuration based on \(r'_k\) is poorly suited to network traffic.

Overall, the main empirical pattern is sensitive to the representation supplied to the temporal head. On BPI2018, the representation \(r'_k\) does not reproduce the clearer recoverability advantage obtained with \(r_k\), while generation quality remains near-equivalent. It also does not reproduce the stronger recoverability or generation quality obtained with \(r_k\) on network traffic. Because the network run with \(r'_k\) uses a different temporal-loss weight, the experiment supports the practical choice of \(r_k\) without isolating the representation as the sole causal factor.

\section{Computational Overhead}
\label{app:compute}
We report the computational cost of the \joint and \twostage training procedures defined in \cref{ss:train_schedules} (\Cref{alg:training-joint,alg:training-twostage}). All experiments were performed on NVIDIA RTX PRO 6000 Blackwell Max-Q GPUs (96\,GB VRAM). The network traffic experiments used two GPUs with a batch size of 32 per GPU (64 total), while the remaining datasets were trained on a single GPU.

Our experiments consider two training strategies. In the proposed \joint training approach, for a single training run of 40 epochs. In \twostage training first trains the language model for 40 epochs and then performs an additional 40 epochs of training using only the auxiliary objective, resulting in a total of 80 training epochs.

\Cref{tab:training-cost} summarizes the training time. For two-stage training, the reported time is separated into the language-model (LM) stage and the auxiliary (Aux) stage.

\begin{table}[t]
\centering
\adjustbox{max width=\columnwidth}{
\begin{tabular}{lccc}
\toprule
Dataset & \joint{} (40 ep.) & \twostage{} (80 ep.) & Peak VRAM \\
\midrule
Network traffic & 58.0 h & 68.7 h & 51 GB \\
BPI2018 & 17 min & 15 min & 3 GB \\
MIMIC-IV & 8.3 h & 9.3 h & 29 GB \\
GDELT & 53.3 min & 66.7 min & 60 GB \\
\bottomrule
\end{tabular}
}
\caption{Total training time and peak GPU memory usage. \joint{} training optimizes both objectives simultaneously for 40 epochs, whereas \twostage{} training consists of 40 epochs of language-model training followed by 40 epochs of auxiliary training.}
\label{tab:training-cost}
\end{table}

The proposed method introduces only a lightweight auxiliary prediction head on top of the language-model backbone. Consequently, the backbone architecture, hidden-state dimension, and sequence length remain unchanged, and the additional computation arises only from evaluating the auxiliary loss during training. Although each epoch of joint training is slightly more expensive than language-model training alone, it eliminates the need for a separate auxiliary optimization stage. As shown in \Cref{tab:training-cost}, this results in comparable or lower end-to-end training time across the evaluated datasets while maintaining a modest memory footprint.

\section{Model Configurations}
\label{app:model-configurations}

\Cref{tab:mamba2-configurations} records the concrete Mamba2 configurations used in our experiments. Following the training schedules defined in \cref{ss:train_schedules} (\Cref{alg:training-joint,alg:training-twostage}), for optimization, all \joint models are trained for 40 epochs.  All \twostage models use 40 epochs of token-only training followed by 40 epochs of temporal-head training with the backbone frozen. For the partially timestamped domains \cref{ss:partially-timestamped}, the temporal head additionally uses four temporal lags.

\parab{Training Hyperparameters and Setup.}
All models are optimized using AdamW \citep{loshchilov2017decoupled} with weight decay $0.1$. We use a base peak learning rate of $\eta = 2 \times 10^{-4}$ (scaled to $4 \times 10^{-4}$ for multi-GPU DDP runs on Network traffic), scheduled with a 1-epoch linear warmup followed by cosine annealing decay down to a floor ratio of $0.1$. Next-token modeling uses Cross-Entropy loss with label smoothing ($\epsilon_{\mathrm{smooth}} = 0.1$), while temporal regression uses Mean Squared Error over log-transformed targets $\log(1 + \Delta t / s)$. Models are trained with a batch size of $64$ for BPI2018, MIMIC-IV, Network traffic (32 per GPU), and GiantMIDI (sequence lengths of $4096$ tokens with $50\%$ window stride, and $256$/$128$ for BPI2018), and a batch size of $32$ for GDELT (sequence length $5000$, stride $1024$).

\begin{table}[t]
  \centering
  \adjustbox{max width=\columnwidth}{%
    \begin{tabular}{lcccccc}
      \toprule
      Dataset & $d_{\mathrm{model}}$ & Layers & Aux head & Hidden dim & $s$ & $\lambda_{\tau}$ \\
      \midrule
      BPI2018 & 128 & 4 & MLP & 256 & 1.0 & 0.1 \\
      MIMIC-IV & 256 & 8 & MLP & 768 & 1.0 & 0.1 \\
      Network traffic & 768 & 24 & GRU & 768 & 0.001 & 300 \\
      GDELT & 256 & 8 & MLP & 768 & 0.001 & 1.0 \\
      GiantMIDI & 192 & 6 & MLP & 384 & 1.0 & 0.1 \\
      \bottomrule
    \end{tabular}%
  }
  \caption{Domain-specific Mamba2 configurations. The column $s$ denotes the
  time scale used in the transformed temporal target, and $\lambda_{\tau}$
  denotes the temporal-loss weight in \cref{eq:joint-objective}.}
  \label{tab:mamba2-configurations}
\end{table}

\section{Domain-Specific Baselines}
\label{app:baseline-details}

This appendix provides the full baseline setup, results, and interpretation summarized in \cref{ss:domain-specific-baselines}, kept separate from the controlled \joint–\twostage comparison in the main text. Here, a \emph{baseline} denotes a previously proposed domain-specific model that serves as an external reference point for the task. Studying such baselines is useful because it helps contextualize the absolute scale of the reported metrics, shows how our models compare with domain-adapted alternatives from prior work, and clarifies which observations appear specific to \modelname{} versus common across different modeling choices. In all cases, the baselines are retrained on the same train/validation/test splits as \modelname{} and evaluated with the same domain-specific metrics. However, they remain descriptive reference points only. Because they differ from \modelname{} in architecture, objective, temporal representation, and generation procedure, they do not isolate the effect of the training regime and therefore cannot be used to answer \textbf{RQ1} (\cref{sec:recoverability}) or \textbf{RQ2} (\cref{sec:results}). The evidence for those two questions comes only from the controlled \joint{}--\twostage{} comparison.

\subsection{Baseline Presentation}

\Cref{tab:baseline-overview} summarizes the baseline family used for each domain, including the event-generation procedure and the temporal mechanism.

\begin{table*}[t]
  \centering
  \setlength{\tabcolsep}{5pt}
  \begin{tabular}{p{2.1cm}p{2.6cm}p{4.6cm}p{4.9cm}}
    \toprule
    Dataset
      & Baseline
      & Event-generation procedure
      & Temporal mechanism \\
    \midrule
    BPI2018
      & LSTM~\cite{tax2017predictive}
      & Autoregressive activity generation until end-of-sequence
      & Next-gap prediction from the shared recurrent state \\
    MIMIC-IV
      & THP~\cite{zuo2020thp}
      & Autoregressive next-event generation from the event-history context
      & Joint next-gap prediction from the same contextual representation \\
    Network traffic
      & NetSSM + GMM~\cite{chu2025netssm}
      & Autoregressive byte-sequence generation with packet delimiters
      & Packet inter-arrival times sampled afterward from a 3-component GMM \\
    Temporal KG
      & RE-Net + GMM~\cite{renet}
      & Autoregressive future-fact generation from temporal quadruples
      & Fact-level temporal gaps sampled afterward from a 3-component GMM \\
    GiantMIDI
      & Music Transformer~\cite{huang2019musictransformer}
      & Autoregressive note-event generation
      & Timing represented through discrete \texttt{time\_shift} tokens \\
    \bottomrule
  \end{tabular}
  \caption{Reference baselines used for descriptive comparison across the four
  main domains and the auxiliary GiantMIDI experiment.}
  \label{tab:baseline-overview}
\end{table*}

\parab{BPI2018.}
For business-process traces, we use an LSTM baseline inspired by the predictive-process-monitoring model of Tax et al.~\cite{tax2017predictive}, which jointly predicts the next activity and its time gap from a recurrent state. In our experiments, we use a modernized implementation rather than the original legacy training script. The model is trained for 50 epochs with batch size 256.

\parab{MIMIC-IV.}
For MIMIC-IV, we use the Transformer Hawkes Process (THP)~\cite{zuo2020thp}, a sequence model designed for event streams with explicit temporal dynamics. We train for 20 epochs with batch size 64, model dimension 256, 4 attention heads, and 4 transformer layers. Relative times are scaled in hours during training.

\parab{Network traffic.}
For network traffic, we use NetSSM~\cite{chu2025netssm} as the content generator and then assign timing with a separately fitted Gaussian mixture model, reflecting a decoupled content/time pipeline. NetSSM is trained on the tokenized packet-sequence representation of the dataset for 10 epochs with batch size 8. Packet timestamps are then assigned using a 3-component GMM fit on training-split inter-arrival times.

\parab{GDELT.}
For temporal knowledge graphs, we use RE-Net~\cite{renet}, again combined with a separate Gaussian mixture model for fact-level timing, similar to what we did with NetSSM. We follow the standard RE-Net training schedule, with a pretraining phase followed by full training; both phases use hidden size 200, dropout 0.5, learning rate $10^{-3}$, batch size 1024, and 20 epochs.

\parab{GiantMIDI.}
For the auxiliary GiantMIDI experiment, we use Music Transformer~\cite{huang2019musictransformer}, a strong symbolic-music baseline in which timing is represented directly through discrete \texttt{time\_shift} tokens. It is trained on PerformanceRNN-style event tokens obtained from the MAESTRO-style export of GiantMIDI~\cite{kong2022giantmidi,hawthorne2019maestro}. We use discrete note and \texttt{time\_shift} events, fix note duration to 0.10 seconds and velocity to 64 during export for 100 epochs with batch size 2.

\begin{table*}[t]
  \centering
  \setlength{\tabcolsep}{4pt}
  \begin{tabular}{lllrrrrr}
    \toprule
    Timestamping
      & Dataset
      & Baseline
      & \multicolumn{5}{c}{Content-generation quality} \\
    \midrule
    \multirow{6}{*}{Dense}
      & \multirow{2}{*}{BPI2018}
      &
      & DLS mean $\uparrow$
      & DLS median $\uparrow$
      &
      &
      & \\
    \cmidrule(lr){3-8}
      &
      & LSTM
      & \textbf{0.535} (0.487)
      & \textbf{0.533} (0.500)
      &
      &
      & \\
    \cmidrule(lr){2-8}
      & \multirow{2}{*}{MIMIC-IV}
      &
      & Token JSD $\downarrow$
      & Recall@50 $\uparrow$
      &
      &
      & \\
    \cmidrule(lr){3-8}
      &
      & THP
      & \textbf{0.111} (0.302)
      & \textbf{1.000} (0.780)
      &
      &
      & \\
    \cmidrule(lr){2-8}
      & \multirow{2}{*}{GiantMIDI}
      &
      & OA mean $\uparrow$
      & Mean pitch OA $\uparrow$
      & Note density OA $\uparrow$
      & Pitch range OA $\uparrow$
      & Pitch var. OA $\uparrow$ \\
    \cmidrule(lr){3-8}
      &
      & Music Transformer
      & \textbf{0.577} (0.172)
      & \textbf{0.458} (0.345)
      & \textbf{0.270} (0.230)
      & \textbf{0.852} (0.050)
      & \textbf{0.730} (0.063) \\
    \midrule
    \multirow{4}{*}{Partial}
      & \multirow{2}{*}{Network traffic}
      &
      & JSD ip.src $\downarrow$
      & JSD ip.dst $\downarrow$
      & JSD tcp.src $\downarrow$
      & JSD tcp.dst $\downarrow$
      & \\
    \cmidrule(lr){3-8}
      &
      & NetSSM
      & 0.3076 (\textbf{0.0412})
      & 0.2426 (\textbf{0.0311})
      & 0.2923 (\textbf{0.0764})
      & 0.2923 (\textbf{0.1120})
      & \\
    \cmidrule(lr){2-8}
      & \multirow{2}{*}{Temporal KG}
      &
      & MRR $\uparrow$
      & MR $\downarrow$
      & Hits@1 $\uparrow$
      & Hits@3 $\uparrow$
      & Hits@10 $\uparrow$ \\
    \cmidrule(lr){3-8}
      &
      & RE-Net
      & \textbf{0.4136} (0.1461)
      & \textbf{144.16} (269.06)
      & \textbf{0.3446} (0.0899)
      & \textbf{0.4398} (0.1490)
      & \textbf{0.5405} (0.2527) \\
    \bottomrule
  \end{tabular}
  \caption{Content-generation results for the reference baselines on the four
  main evaluation domains and the auxiliary GiantMIDI experiment. Parenthesized
  values report the best ChronoSSM result across \joint{} and \twostage{};
  boldface marks the better of the baseline and ChronoSSM value.}
  \label{tab:baseline-generation}
\end{table*}

\subsection{Baseline Recoverability Protocol}

The recoverability analysis requires a small adaptation for the external baselines, because they do not all expose the same temporal representation as \modelname{}. The controlled \joint{}--\twostage{} comparison always probes the representation supplied to the temporal head. For the baselines, we therefore probe the frozen representation most directly tied to temporal prediction in each model, so that the diagnostic remains as comparable as possible across methods. This is the recurrent state for the LSTM, the event-history representation for THP, the packet-boundary representation for NetSSM, and the fact-history representation for RE-Net. Within each dataset, the baseline probe uses the same target \(z_k\) and the same evaluation protocol as the controlled \joint{}--\twostage{} comparison. The resulting numbers remain descriptive only, since the compared models still differ in architecture and temporal parameterization. \Cref{tab:baseline-recoverability} reports the resulting descriptive recoverability scores.

\begin{table*}[t]
  \centering
  \begin{tabular}{lllrrr}
  \toprule
  Timestamping
    & Dataset
    & Baseline
    & TCS $\uparrow$
    & Probe \(R^2\) $\uparrow$
    & Probe MAE $\downarrow$ \\
  \midrule
  \multirow{3}{*}{Dense}
    & BPI2018
    & LSTM
    & 0.0375 (\textbf{0.8030})
    & 0.0001 (\textbf{0.631})
    & 4.120  (\textbf{2.775}) \\
  \cmidrule(lr){2-6}
    & MIMIC-IV
    & THP
    & 0.8584 (\textbf{0.8620})
    & 0.2265 (\textbf{0.6580})
    & 1.96 (\textbf{0.5970}) \\
  \cmidrule(lr){2-6}
    & GiantMIDI
    & Music Transformer
    & -1.000 (\textbf{-0.044})
    & undef. (\textbf{0.160})
    & \textbf{10.568} (248.006) \\
  \midrule
  \multirow{2}{*}{Partial}
    & Network traffic
    & NetSSM + GMM
    & 0.3689 (\textbf{0.6216})
    & -0.5744 (\textbf{0.3943})
    & 0.860 (\textbf{0.3080}) \\
  \cmidrule(lr){2-6}
    & Temporal KG
    & RE-Net + GMM
    & \textbf{0.3762} (0.2027)
    & 0.0161 (\textbf{0.4449})
    & 1,412.25 (\textbf{904.14}) \\
  \bottomrule
  \end{tabular}
  \caption{Temporal-recoverability results for the reference baselines on the
  four main evaluation domains and the auxiliary GiantMIDI experiment.
  Parenthesized values report the best ChronoSSM result across \joint{} and
  \twostage{}; boldface marks the better of the baseline and best ChronoSSM
  value.}
  \label{tab:baseline-recoverability}
\end{table*}

\subsection{Baseline Results}

The baseline comparisons are most informative when they reveal a separation between generation quality and temporal recoverability. On several datasets, the domain-specific baselines are clearly stronger than \modelname{} on the generation task, which is expected because they were designed specifically for that domain. \Cref{tab:baseline-generation} shows that, on BPI2018, the LSTM yields higher suffix similarity than both controlled variants. On MIMIC-IV, THP also dominates the generation metrics. The largest gap appears on the temporal knowledge graph, where RE-Net substantially outperforms \modelname{} on filtered link prediction. The auxiliary GiantMIDI baseline shows the same pattern, with Music Transformer far ahead on OA generation quality. These gaps are not a problem for the main claims of the paper; they are expected, because these baselines are purpose-built generators for their respective domains.

By contrast, the recoverability picture is consistently weak or at least much less convincing for the baselines. The LSTM baseline on BPI2018 has almost no recoverability under the probe diagnostics. NetSSM is similarly weak on network traffic, including a negative probe \(R^2\). RE-Net is another important case: despite much better generation quality, its probe recoverability remains poor relative to both ChronoSSM variants. Music Transformer also fits this pattern, with very poor recoverability despite being the strongest generator in that setting.

The most revealing case is THP on MIMIC-IV. Its TCS is close to that of \joint{}, yet its probe \(R^2\) is much lower. This suggests that a seemingly favorable TCS value does not by itself imply that temporal information is cleanly or linearly accessible from the representation. More broadly, these baselines do not learn timing through the same kind of shared representation studied in \modelname{}: some use different temporal parameterizations, and others decouple timing from content generation entirely. Their recoverability is therefore generally weak, and when a baseline comes close to \joint{} on a single metric, we interpret that locally as a reminder that no single recoverability metric should be read in isolation.

\section{Sensitivity to Distance Metric and Neighborhood Size}
\label{app:metric-sensitivity}

The Temporal Cohesion Score (TCS) depends on two design choices:  
(i) the distance metric used to compare hidden states, and  
(ii) the neighborhood size $K$ used to define local neighborhoods.

In the main experiments (\cref{sec:recoverability}), we fix $K=5$ and use cosine similarity. In this section, we analyze the robustness of TCS to these choices and clarify how the resulting sensitivity differs between the network-traffic and temporal-knowledge-graph domains.

\subsection{Sensitivity to Distance Metric}
\label{app:metric-sensitivity-network}

We first examine the effect of the distance metric while keeping the neighborhood size fixed at $K=5$. Specifically, we recompute TCS using Euclidean ($L_2$) and Manhattan ($L_1$) distances in place of cosine similarity.

\Cref{tab:tcs-robustness} reports the resulting scores for both domains. On network traffic, the \joint{} model consistently achieves a TCS of approximately $0.62$, while the Two-Stage baseline remains around $0.15$. The relative improvement induced by joint training is therefore large and stable across all three metrics. This robustness indicates that the strong temporal signal observed in network traffic is not an artifact of a particular geometric choice.

On the temporal knowledge graph, the same qualitative pattern holds, but the gains are much smaller: the Two-Stage baseline ranges from $0.104$ to $0.112$, whereas the \joint{} model ranges from $0.126$ to $0.144$. Temporal cohesion is therefore somewhat sensitive to the distance metric in this domain, but the advantage of \joint{} training remains positive under all three choices.

\begin{table}[H]
  \centering
  \adjustbox{max width=\columnwidth}{%
    \begin{tabular}{llrrr}
    \toprule
    & & \multicolumn{3}{c}{TCS Metric Formulation $\uparrow$} \\
    \cmidrule(lr){3-5}
    Dataset & Model & Cosine & Euclidean ($L_2$) & Manhattan ($L_1$) \\
    \midrule
    \multirow{3}{*}{Network Traffic} 
      & \twostage{} & 0.1469 & 0.1542 & 0.1526 \\
      & \textbf{\joint{}} & \textbf{0.6214} & \textbf{0.6225} & \textbf{0.6255} \\
      \cmidrule{2-5}
      & \textit{Improv.} & \textit{+323\%} & \textit{+303\%} & \textit{+310\%} \\
    \midrule
    \multirow{3}{*}{Temporal KG}
      & \twostage{} & 0.110 & 0.112 & 0.104 \\
      & \textbf{\joint{}} & \textbf{0.144} & \textbf{0.126} & \textbf{0.127} \\
      \cmidrule{2-5}
      & \textit{Improv.} & \textit{+31\%} & \textit{+12\%} & \textit{+22\%} \\
    \bottomrule
    \end{tabular}%
  }
  \caption{TCS across distance metrics for network traffic and temporal
  knowledge graphs.}
  \label{tab:tcs-robustness}
\end{table}

\subsection{Sensitivity to Neighborhood Size $K$}

We next vary neighborhood size $K$ on the two partially timestamped domains, again using cosine, Euclidean, and Manhattan distances. \Cref{tab:k-sensitivity} shows a clear contrast between them. For network traffic, the \joint{} model remains far above the Two-Stage baseline for all tested values of $K$ and all three distance metrics, even though TCS gradually decreases as neighborhoods grow. This indicates that the temporal organization induced by \joint{} training is not limited to a narrow nearest-neighbor effect.

For temporal knowledge graphs, the picture is different. The advantage of \joint{} training is visible at small neighborhoods, but it rapidly shrinks toward zero as $K$ increases, regardless of the distance metric.
Here, temporal information is therefore much more localized: once the neighborhood expands, static relational structure dominates similarity and temporal ordering becomes difficult to recover. Together with the distance-metric analysis above, this comparison shows that the main conclusion is robust to the exact geometric choice, while the spatial extent of temporal organization depends strongly on the domain.

\begin{table}[H]
  \centering
  \adjustbox{max width=\columnwidth}{%
    \begin{tabular}{llrrrrrr}
    \toprule
    & & \multicolumn{2}{c}{\textbf{Cosine Distance}} &
      \multicolumn{2}{c}{\textbf{Euclidean ($L_2$)}} &
      \multicolumn{2}{c}{\textbf{Manhattan ($L_1$)}} \\
    \cmidrule(lr){3-4} \cmidrule(lr){5-6} \cmidrule(lr){7-8}
    Dataset & $K$ & \twostage{} & \joint{} &
      \twostage{} & \joint{}&
      \twostage{} & \joint{} \\
    \midrule
    \multirow{4}{*}{Network Traffic}
      & $5$  & 0.147 & \textbf{0.621} & 0.154 & \textbf{0.622} & 0.153 & \textbf{0.626} \\
      & $15$ & 0.032 & \textbf{0.527} & 0.046 & \textbf{0.531} & 0.049 & \textbf{0.533} \\
      & $25$ & 0.049 & \textbf{0.479} & 0.066 & \textbf{0.486} & 0.067 & \textbf{0.488} \\
      & $35$ & 0.061 & \textbf{0.446} & 0.077 & \textbf{0.454} & 0.080 & \textbf{0.456} \\
    \midrule
    \multirow{4}{*}{Temporal KG}
      & $5$  & 0.172 & \textbf{0.203} & 0.112 & \textbf{0.126} & 0.104 & \textbf{0.127} \\
      & $15$ & 0.048 & \textbf{0.051} & 0.038 & \textbf{0.049} & 0.041 & \textbf{0.049} \\
      & $25$ & 0.020 & \textbf{0.029} & 0.029 & \textbf{0.029} & 0.026 & \textbf{0.029} \\
      & $35$ & -0.001 & \textbf{0.000} & 0.001 & \textbf{0.002} & -0.001 & \textbf{0.002} \\
    \bottomrule
    \end{tabular}%
  }
  \caption{Sensitivity of TCS to neighborhood size $K$ for network traffic and
  temporal knowledge graphs.}
  \label{tab:k-sensitivity}
\end{table}

\section{Layer-Wise Representation Analysis}
\label{app:representation-analysis}

This appendix examines how temporal supervision changes the internal representations learned by \modelname{}. It is intended to complement the recoverability results of \cref{sec:recoverability} and the generation-quality results of \cref{sec:results}, rather than to provide a complete mechanistic account of all four evaluation domains.

\subsection{Representation-Similarity Metrics}
\label{app:representation_similarity}

This subsection provides the formal definitions of the representation-similarity metrics used in the layer-wise analysis below. These metrics quantify alignment between hidden representations learned under \joint and \twostage training at different geometric scales.

To quantify structural differences between the two training regimes, we compare hidden states produced by corresponding backbone layers on identical input sequences.
Let $\mathbf{X} \in \mathbb{R}^{n \times d}$ and $\mathbf{Y} \in \mathbb{R}^{n \times d}$
denote the activations from the \joint and \twostage models, respectively, for $n$ samples and representation dimension $d$.
We measure alignment at three complementary geometric scales.

\parab{Global structural alignment (CKA).}
We use linear Centered Kernel Alignment (CKA)~\cite{pmlr-v97-kornblith19a} to measure similarity between the global ``clouds'' formed by the representations. Given Gram matrices $K = \mathbf{X}\mathbf{X}^\top$ and $L = \mathbf{Y}\mathbf{Y}^\top$, and the centering matrix $H = I - \frac{1}{n}\mathbf{1}\mathbf{1}^\top$, linear CKA is defined as
\begin{equation}
\text{CKA}(K, L) =
\frac{\text{HSIC}(K, L)}
{\sqrt{\text{HSIC}(K, K)\,\text{HSIC}(L, L)}},
\end{equation}
where
\begin{equation}
\text{HSIC}(K, L) = \frac{1}{(n-1)^2}\,\mathrm{tr}(KHLH)
\end{equation}
is the Hilbert--Schmidt Independence Criterion~\cite{NIPS2007_d5cfead9}. High CKA values indicate strong global alignment between representation spaces.

\parab{Local neighborhood alignment (CKNNA).}
To assess alignment at a local geometric scale, we use Centered Kernel Nearest Neighbor Alignment (CKNNA)~\cite{pmlr-v235-huh24a}.
Let $\mathbf{A}$ be a binary adjacency matrix such that $\mathbf{A}_{ij}=1$ if sample $j$ is a mutual $k$-nearest neighbor of sample $i$ in both representation spaces. CKNNA is defined as
\begin{equation}
\text{CKNNA}(K, L) =
\frac{\mathrm{tr}\!\left((\mathbf{A} \circ K)(\mathbf{A} \circ L)^\top\right)}
{\|\mathbf{A} \circ K\|_F \cdot \|\mathbf{A} \circ L\|_F},
\end{equation}
where $\circ$ denotes the Hadamard product and $\|\cdot\|_F$ is the Frobenius norm. Lower values indicate greater divergence in local neighborhood structure between the two representations.

\parab{Topological overlap (M-KNN).}
Finally, we compute the Mutual $k$-Nearest Neighbor (M-KNN) overlap to measure agreement in neighborhood membership.
Let $\mathcal{N}_k^{\mathbf{X}}(i)$ denote the set of $k$ nearest neighbors of sample $i$ in representation space $\mathbf{X}$. The M-KNN overlap is defined as
\begin{equation}
\text{M-KNN}(\mathbf{X}, \mathbf{Y}) =
\frac{1}{nk} \sum_{i=1}^n
\left| \mathcal{N}_k^{\mathbf{X}}(i) \cap \mathcal{N}_k^{\mathbf{Y}}(i) \right|.
\end{equation}
Lower overlap indicates that the two models induce different local topological structures over the data.

\begin{figure*}[t!]
\centering
\begin{subfigure}[b]{0.48\textwidth}
\centering
\includegraphics[width=\textwidth]
{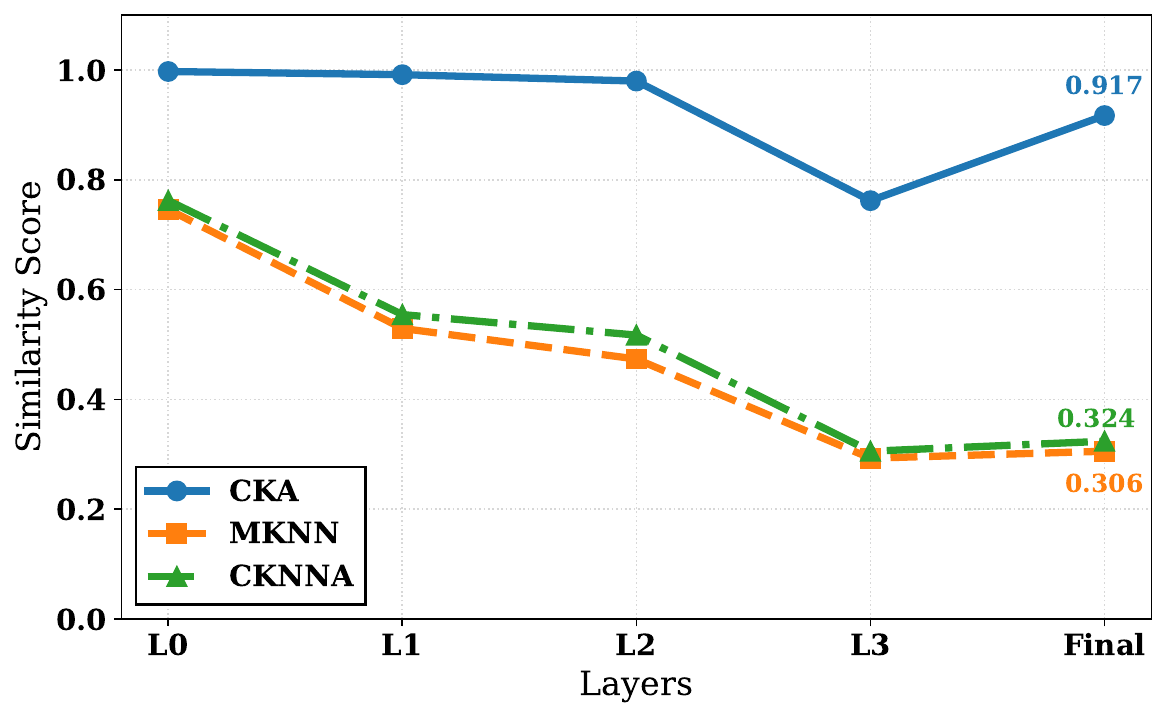}
\caption{BPI2018.}
\label{fig:alignment-bpi}
\end{subfigure}
\hfill
\begin{subfigure}[b]{0.48\textwidth}
\centering
\includegraphics[width=\textwidth]
{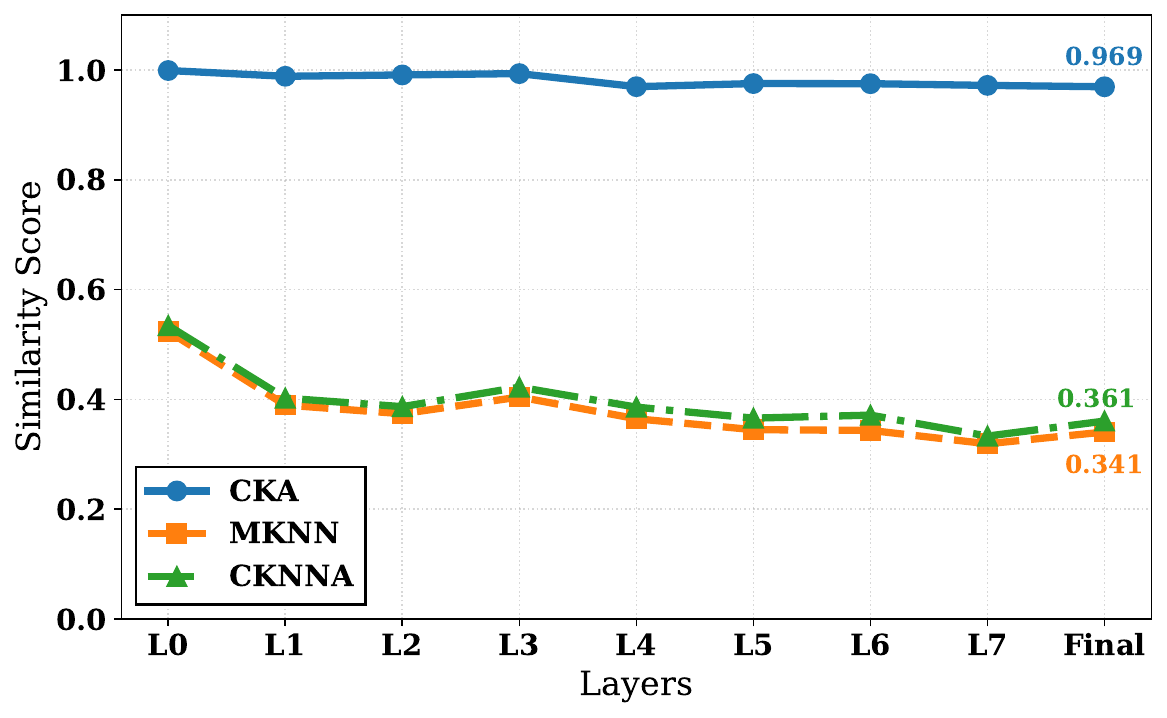}
\caption{MIMIC-IV.}
\label{fig:alignment-mimic}
\end{subfigure}

\medskip

\begin{subfigure}[b]{0.48\textwidth}
\centering
\includegraphics[width=\textwidth]
{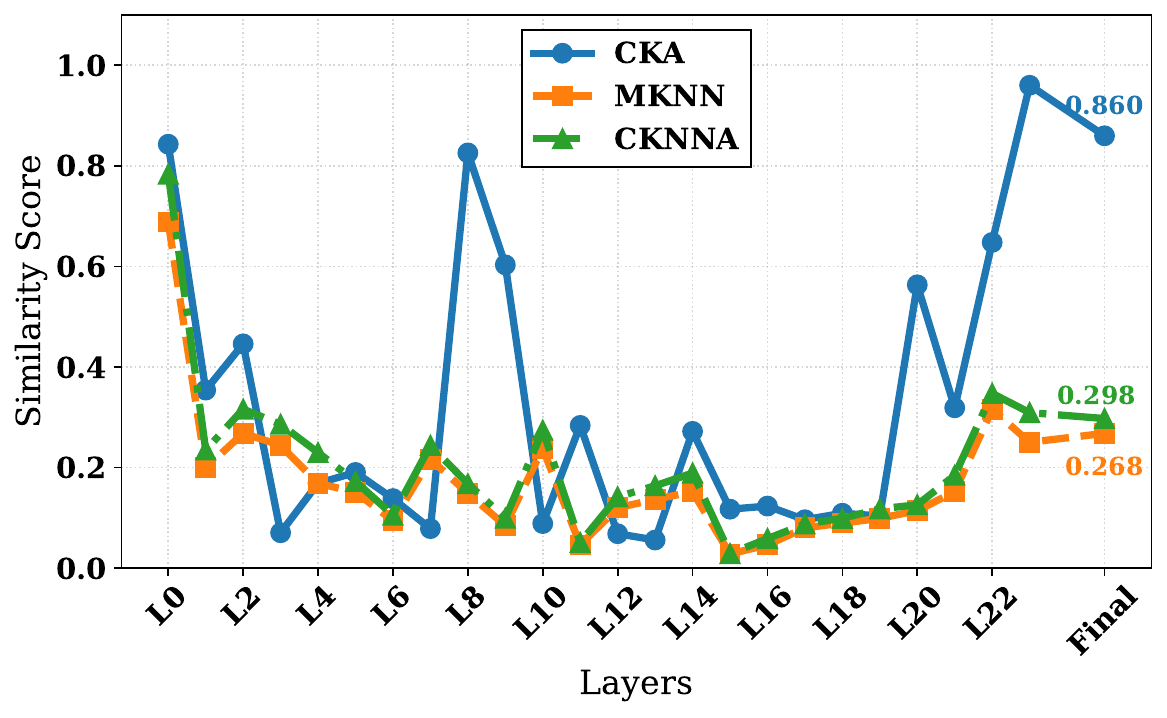}
\caption{Network traffic.}
\label{fig:alignment-network}
\end{subfigure}
\hfill
\begin{subfigure}[b]{0.48\textwidth}
\centering
\includegraphics[width=\textwidth]
{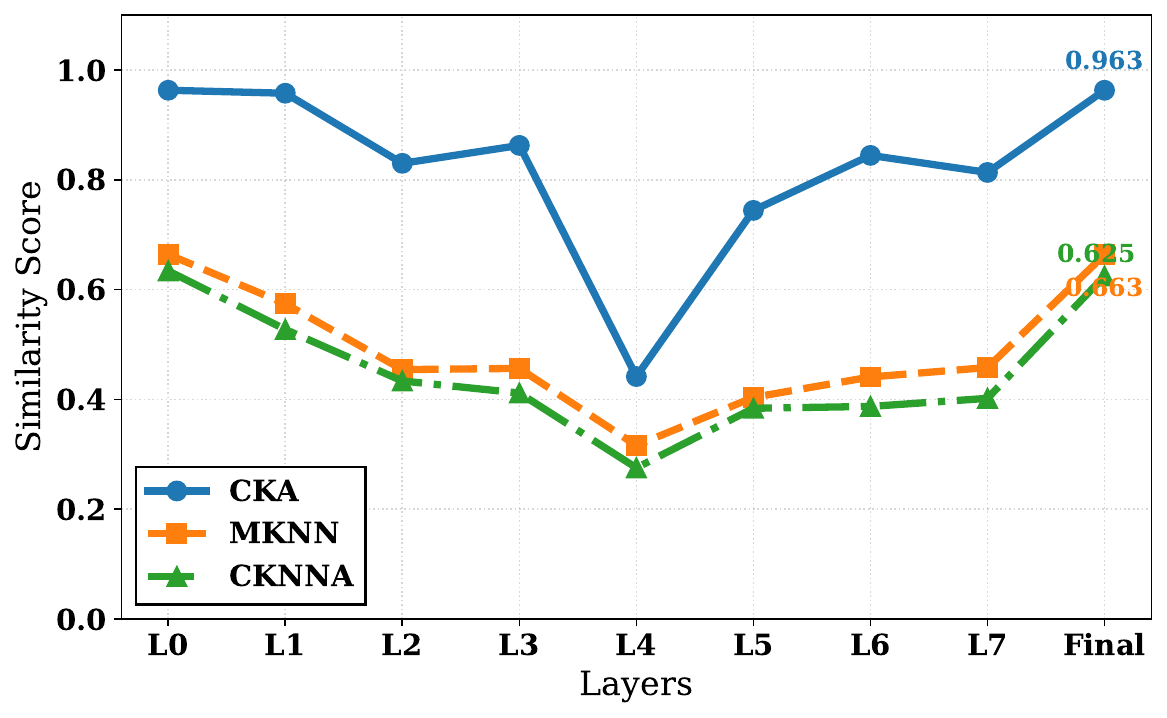}
\caption{Temporal knowledge graph.}
\label{fig:alignment-dkg}
\end{subfigure}
\caption{Layer-wise alignment between \modelname{} models trained under
\joint{} and \twostage{} regimes, measured using CKA, M-KNN, and CKNNA.
CKA captures global geometric alignment, whereas M-KNN and CKNNA emphasize
local neighborhood structure.}
\label{fig:layerwise-alignment}
\end{figure*}

\subsection{Per-Domain Patterns}

\Cref{fig:layerwise-alignment} collects the layer-wise alignment plots
for all four domains.

\parab{BPI2018.}
\Cref{fig:alignment-bpi} shows the mildest change of the four domains. CKA remains high throughout the backbone and ends at \(0.917\), while M-KNN and CKNNA decline but stay well above the network-traffic values. The substantial probe-based recoverability gain despite this relatively mild geometric change is consistent with the possibility that \joint{} makes temporal information more linearly accessible without a major reshaping of the representation space. The alignment measures alone do not establish that mechanism.

\parab{MIMIC-IV.}
\Cref{fig:alignment-mimic} shows a different pattern: CKA stays close to \(1\) across all layers and ends at \(0.969\), whereas M-KNN and CKNNA drop to roughly \(0.34\) to \(0.36\) at the final layer. The natural reading is that \joint{} changes local neighborhood structure much more than global geometry. This local change is compatible with the favorable probe diagnostics on MIMIC-IV, but the plot does not establish how it produces them or explain why generation quality is slightly mixed in this domain.

\parab{Network traffic.}
\Cref{fig:alignment-network} shows the strongest reorganization. CKA is high at the input, drops sharply through much of the backbone, and recovers only near the end, while M-KNN and CKNNA remain low across most layers. This pattern is consistent with the concurrent gains in TCS and probe diagnostics: compared with \twostage{}, \joint{} appears to reshape fine-grained local geometry much more substantially in this domain than in the others. It may therefore be compatible with changes in both local temporal coherence and linear accessibility, without establishing a causal explanation for either the recoverability or generation results.

\parab{Temporal knowledge graphs.}
\Cref{fig:alignment-dkg} lies between the previous two cases. The two models stay well aligned early, diverge most clearly in the middle layers, and recover substantial similarity near the output. This fits the smaller recoverability differences relative to the other domains, but does not provide a clear account of why those differences are small: \joint{} changes the intermediate computation, while the two regimes converge again toward more similar late-layer geometries.

\subsection{Cross-Domain Interpretation}
\label{appsec:cross-domain-inter}
The main cross-domain observation is negative but informative: the four domains do not support a single geometric story for why \joint{} improves temporal recoverability. BPI2018 shows only modest change, MIMIC-IV preserves almost the same global geometry while altering local neighborhoods, temporal knowledge graphs diverge mainly in the middle layers before recovering, and network traffic undergoes the broadest local and intermediate-layer reorganization.

This variation weakens two simple explanations at once. First, it weakens a single-mechanism interpretation in which improved recoverability would always arise from the same kind of representational change. Second, it weakens a simple dense-versus-partial timestamping explanation: BPI2018 and MIMIC-IV are both densely timestamped but behave quite differently, and the two partially timestamped domains are likewise not interchangeable. Timestamp density alone is therefore too coarse to explain the observed patterns.

There is no simple relationship between alignment and recoverability gains. BPI2018 exhibits a substantial probe-based gain despite relatively high final-layer alignment, whereas network traffic combines broad local divergence with large gains in both TCS and the probe diagnostics. These observations are consistent with the domain-specific interpretations in the recoverability analysis, but the plots do not verify them or identify a common mechanism.

The safest conclusion is therefore narrow. \joint{} can improve temporal recoverability under several different patterns of representational change, and those patterns do not map cleanly onto downstream generation quality. The layer-wise analysis is useful mainly because it rules out overly simple explanations; it does not provide a complete account of why the domain-level results differ. This interpretation is robust to the neighborhood size used for M-KNN and CKNNA: although the absolute values change, the same qualitative patterns remain.
\end{document}